\documentclass{article}

\usepackage[preprint]{neurips_2026}

\usepackage[utf8]{inputenc} 
\usepackage[T1]{fontenc}    
\usepackage{hyperref}       
\usepackage{url}            
\usepackage{booktabs}       
\usepackage{amsfonts}       
\usepackage{nicefrac}       
\usepackage{microtype}      
\usepackage{xcolor}         

\usepackage{enumitem}
\usepackage{amssymb}
\usepackage{amsmath}
\usepackage{graphicx}
\usepackage{multirow}
\usepackage{colortbl}
\usepackage{wrapfig}

\title{\textsc{DiffScore}: \\ Text Evaluation Beyond Autoregressive Likelihood}

\author{
\textbf{
Wen Lai$^{1}$\thanks{Equal contribution.}\hspace{0.08cm} \thanks{Work mostly done while affiliated with the Technical University of Munich.} \quad
Yingli Shen$^{2*}$ \quad
Dingnan Jin$^{1}$ \quad
Qing Cui$^{1}$ \quad
Jun Zhou$^{1}$ \quad
} \\
\textbf{
Maosong Sun$^{2}$ \quad
Alexander Fraser$^{3}$ \quad
}
\\ \\
$^1$Ant Group \quad
$^2$Tsinghua University \quad
$^3$Technical University of Munich \\
\texttt{wen.lai@tum.de, syl@mail.tsinghua.edu.cn}
}

\begin{document}

\maketitle

\begin{abstract}
Autoregressive language models are widely used for text evaluation, however, their left-to-right factorization introduces positional bias, i.e., early tokens are scored with only leftward context, conflating architectural asymmetry with true text quality.
We propose \emph{masked reconstruction} as an alternative paradigm, where every token is scored using full bidirectional context.
We introduce \textsc{DiffScore}, an evaluation framework built on Masked Large Diffusion Language Models.
By measuring text recoverability across continuous masking rates, \textsc{DiffScore} eliminates positional bias and naturally establishes an evaluation hierarchy from local fluency to global coherence.
We further provide diagnostic tools unavailable to autoregressive frameworks: \emph{multi-timestep quality profiles} that decompose scores across masking rates, and \emph{bidirectional PMI decomposition} that disentangles fluency from faithfulness.
Experiments across ten benchmarks show that \textsc{DiffScore} consistently outperforms autoregressive baselines in both zero-shot and fine-tuned settings.
The code is released at: \url{https://github.com/wenlai-lavine/DiffScore}.
\end{abstract}

\vspace{-1.5em}
\begin{figure}[h]
    \centering
    \includegraphics[width=0.95\linewidth]{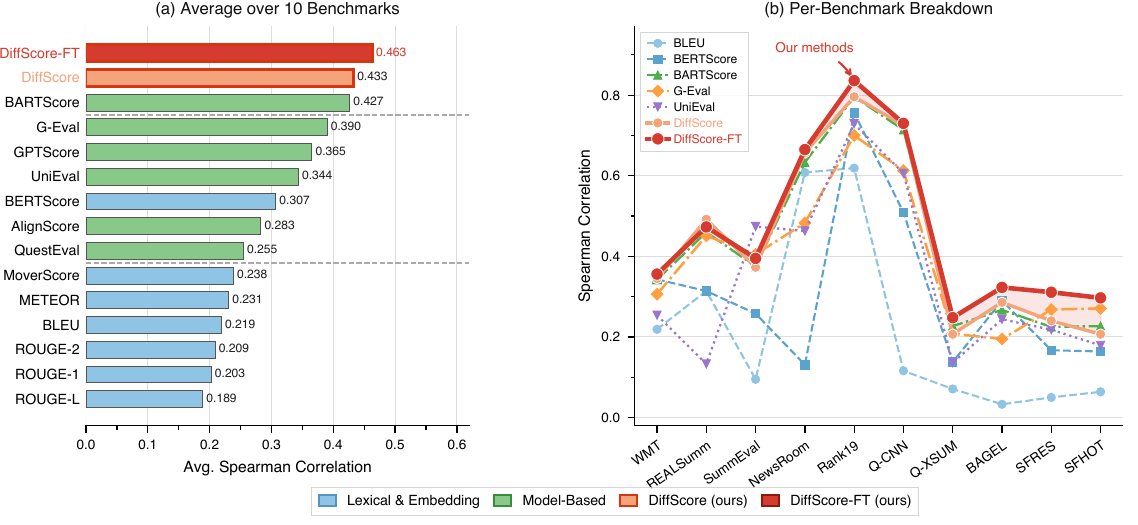}
    \vspace{-1em}
    \caption{\textsc{DiffScore} consistently outperforms all baselines across 10 diverse evaluation benchmarks.}
    \label{fig:radar_main}
\end{figure}

\section{Introduction}
\label{sec:intro}

Evaluating natural language generation (NLG) remains challenging due to substantial lexical variation among semantically equivalent outputs~\cite{clark-etal-2021-thats, gehrmann2023repairing}.
Methods have evolved from $n$-gram overlap (BLEU~\cite{papineni-etal-2002-bleu}, ROUGE~\cite{lin-2004-rouge}) and semantic matching (BERTScore~\cite{Zhang*2020BERTScore:}, MoverScore~\cite{zhao-etal-2019-moverscore}) to generative scoring (BARTScore~\cite{yuan2021bartscore}, GPTScore~\cite{fu-etal-2024-gptscore}) and LLM-as-Judge approaches (G-Eval~\cite{liu-etal-2023-g}).
Among these, autoregressive (AR) scoring via conditional log-likelihood $\sum_n \log p(x^n \mid x^{<n}, \mathbf{s})$ remains a principled framework.
However, AR scoring carries a fundamental limitation: left-to-right factorization conditions each token solely on preceding context, introducing \emph{positional bias} whereby early tokens are evaluated under information-impoverished contexts with lower signal-to-noise ratios (\S\ref{ssec:pos_dir}).
The Reversal Curse~\cite{berglund2024the}, whereby AR models fail to infer ``B is A'' from ``A is B,'' further exposes an intrinsic directional asymmetry, suggesting AR scores reflect architectural bias as much as true text quality.

We propose modeling text evaluation as \emph{masked reconstruction}: measuring a model's ability to recover candidate text under varying degrees of context ablation.
High-quality text conforms to linguistic regularities and is thus confidently reconstructable under arbitrary masking patterns.
Masked Large Diffusion Language Models (MDLLMs;~\cite{sahoo2024simple, nie2025large,ye2025dream}) provide an ideal backbone, offering three structural advantages:
(1)~\textbf{Bidirectionality} eliminates positional bias by conditioning on bilateral context;
(2)~\textbf{Multi-granularity signals} from diverse masking rates span an evaluation hierarchy from local fluency to global coherence; and
(3)~\textbf{Native objective alignment} ensures the pretraining objective directly matches the evaluation task.

We instantiate this paradigm as \textsc{DiffScore}, the first masked reconstruction-based NLG evaluation framework, in two variants.
\textbf{DiffScore-Zero} performs zero-shot quality estimation via the MDLLM's evidence lower bound (ELBO).
\textbf{DiffScore-FT} is fine-tuned on task-relevant corpora without gold-standard references, isolating the performance contribution of the evaluation paradigm itself.
For interpretability, we introduce two diagnostic tools:
(1)~\emph{multi-timestep quality profiles}, which decompose scores across masking rates; and
(2)~\emph{bidirectional PMI decomposition}, which disentangles fluency and faithfulness into independent dimensions, a capability structurally unattainable in AR frameworks.

Experiments across 10 benchmarks spanning summarization, machine translation, and data-to-text generation validate our approach.
As shown in Figure~\ref{fig:radar_main}, DiffScore-FT outperforms BARTScore in most settings, while DiffScore-Zero surpasses GPTScore under zero-shot conditions, highlighting MDLLMs as potent out-of-the-box evaluators.
Analyses confirm lower positional bias, stronger directional consistency, and superior interpretability relative to AR baselines.

Our contributions are:
\textbf{(I)}~We introduce masked reconstruction for NLG evaluation and build \textsc{DiffScore}, the first MDLLM-based evaluation framework, directly addressing the structural flaws of AR scoring.
\textbf{(II)}~We propose multi-timestep quality profiles and bidirectional PMI decomposition, enabling multi-dimensional quality insights beyond AR frameworks.
 \textbf{(III)}~We demonstrate that \textsc{DiffScore}, built on open-weight MDLLMs, offers a competitive and fully reproducible alternative to proprietary LLM-based evaluators such as G-Eval~\citep{liu-etal-2023-g}, requiring neither API access nor prompt engineering.
\section{Preliminaries}
\label{sec:preliminary}

We establish notation and review the two foundations of our framework: autoregressive text evaluation (\S\ref{ssec:ar_eval}) and masked diffusion language models (\S\ref{ssec:mdlm}).

\noindent\textbf{Notation.}
Let $\mathcal{V}$ be a finite vocabulary, $\texttt{[M]} \notin \mathcal{V}$ a mask token, and $\bar{\mathcal{V}} \triangleq \mathcal{V} \cup \{\texttt{[M]}\}$. A length-$L$ sequence is denoted $\mathbf{x} = (x^1, \ldots, x^L) \in \mathcal{V}^L$. For evaluation, we assess a \emph{candidate} text $\mathbf{c} = (c^1, \ldots, c^{L_c})$ given a \emph{source} text $\mathbf{s} = (s^1, \ldots, s^{L_s})$.

\subsection{Autoregressive Text Evaluation}
\label{ssec:ar_eval}

An autoregressive (AR) language model factorizes sequence log-probability left-to-right:
\begin{equation}
  \label{eq:ar}
  \log p_{\mathrm{AR}}(\mathbf{x})
    = \sum_{n=1}^{L} \log p_{\mathrm{AR}}(x^n \mid x^{<n}).
\end{equation}
\textsc{BARTScore}~\cite{yuan2021bartscore} repurposes this for text evaluation, scoring a candidate $\mathbf{c}$ given a source $\mathbf{s}$ via:
\begin{equation}
  \label{eq:bartscore}
  \textsc{BARTScore}(\mathbf{c} \mid \mathbf{s})
    = \sum_{n=1}^{L_c}
      \log p_{\mathrm{AR}}(c^n \mid c^{<n},\, \mathbf{s}),
\end{equation}
where $p_{\mathrm{AR}}$ is a task-finetuned BART-large~\cite{lewis-etal-2020-bart}. Three variants capture distinct quality dimensions: \emph{marginal} $\textsc{BARTScore}(\mathbf{c})$ for intrinsic fluency; \emph{reverse} $\textsc{BARTScore}(\mathbf{s} \mid \mathbf{c})$ for information coverage; and \emph{bidirectional} (averaging both) for holistic assessment.
However, AR factorization introduces two structural flaws~\cite{berglund2024the}: \emph{positional bias}, as early tokens are evaluated with impoverished context due to unidirectional conditioning; and \emph{directional asymmetry}, where the fixed order conflates architectural artifacts with intrinsic text quality.

\subsection{Masked Large Diffusion Language Models}
\label{ssec:mdlm}

Masked large diffusion language models~\cite{sahoo2024simple,nie2025large,ye2025dream} define a discrete diffusion process over sequences using $\texttt{[M]}$ as an absorbing state.

\noindent\textbf{Forward (masking) process.}
At continuous time $t \in [0,1]$, the forward process independently corrupts each token of a clean sequence $\mathbf{x}_0 \in \mathcal{V}^L$:
\begin{equation}
  \label{eq:forward}
  q(x_t^i \mid x_0^i)
    = (1 - t)\,\mathbb{1}[x_t^i = x_0^i]
    \;+\; t\,\mathbb{1}[x_t^i = \texttt{[M]}],
\end{equation}
replacing it with $\texttt{[M]}$ with probability $t$. Let $\mathbf{x}_t \in \bar{\mathcal{V}}^L$ denote the corrupted sequence and $\mathcal{M}_t \triangleq \{i : x_t^i = \texttt{[M]}\}$ the masked positions. The sequence transitions from fully intact ($t{=}0$) to fully masked ($t{=}1$).

\noindent\textbf{Reverse (denoising) process.}
A neural network $\theta$ predicts clean tokens at masked positions conditioned on the partially masked sequence:
\begin{equation}
  \label{eq:reverse}
  p_\theta(\hat{\mathbf{x}}_0 \mid \mathbf{x}_t)
    = \prod_{i \in \mathcal{M}_t}
      p_\theta(\hat{x}_0^i \mid \mathbf{x}_t).
\end{equation}
Crucially, $p_\theta$ is a bidirectional Transformer \emph{without causal masking}, so each prediction attends to all unmasked tokens regardless of position, unlike AR unidirectional conditioning (Eq.~\ref{eq:ar}).

\noindent\textbf{Evidence lower bound.}
The log-marginal likelihood admits a variational lower bound~\cite{sahoo2024simple}:
\begin{equation}
  \label{eq:elbo}
  \log p_\theta(\mathbf{x}_0)
    \;\geq\;
    \underbrace{
      \mathbb{E}_{t \sim \mathcal{U}(0,1)}\;
      \mathbb{E}_{\mathbf{x}_t \sim q(\mathbf{x}_t \mid \mathbf{x}_0)}
      \!\left[\,
        \frac{1}{t}
        \sum_{i \in \mathcal{M}_t}
          \log p_\theta(x_0^i \mid \mathbf{x}_t)
      \,\right]
    }_{\displaystyle
      \triangleq\;\mathrm{ELBO}(\mathbf{x}_0;\,\theta)}.
\end{equation}
The $1/t$ factor reweights the expected number of masked tokens ($\mathbb{E}[|\mathcal{M}_t|] \approx tL$), equalizing per-token contributions across timesteps. The ELBO is estimated by uniformly sampling $K$ timesteps $t_k \sim \mathcal{U}(0,1]$ and masking patterns $\mathbf{x}_{t_k} \sim q(\mathbf{x}_{t_k} \mid \mathbf{x}_0)$.
\section{\textsc{DiffScore}}
\label{sec:diffscore}

We conceptualize text evaluation as measuring \emph{reconstruction fidelity}: high-quality text should be robustly recoverable from partial masking. Operationalizing this via the ELBO of MDLLMs endows \textsc{DiffScore} with two structural advantages over AR metrics: (1) \textbf{Bidirectionality}, which eliminates positional bias; and (2) \textbf{Factorization-invariance}, as it marginalizes over random masking patterns rather than relying on a deterministic generation order. A formal analysis is provided in Appendix~\ref{app:theory}.

\subsection{Scoring Configurations and Estimation}
\label{ssec:scoring_configs}

To evaluate a candidate $\mathbf{c}$ (length $L_c$) given a source $\mathbf{s}$ (length $L_s$), we uniformly sample $K$ masking timesteps $t_k \sim \mathcal{U}(0,1]$ and independent masking patterns $\mathbf{x}_{t_k} \sim q(\mathbf{x}_{t_k} \mid \mathbf{x})$. Let $\mathcal{M}_{t_k}$ denote the masked positions. We define a generalized empirical estimator:
\begin{equation}
  \label{eq:estimator}
  \mathcal{S}(\mathbf{y} \mid \mathbf{x}) = \frac{1}{|\mathbf{y}|} \frac{1}{K} \sum_{k=1}^{K} \omega(t_k) \sum_{i \in \mathcal{M}_{t_k}} \log p_\theta(y^i \mid \mathbf{x}, \mathbf{y}_{t_k}),
\end{equation}
where $\omega(t_k)$ is a weighting function. We default to the \emph{mean log-probability} (\textsc{mlp}) variant where $\omega(t_k) = 1/|\mathcal{M}_{t_k}|$, which yields interpretable per-token scores and avoids numerical instability at low masking rates compared to the strict ELBO weighting $\omega(t_k) = 1/t_k$. Since MDLLMs evaluate all masked positions independently, the $K$ forward passes are fully parallelizable.

Mirroring directional variants in prior work~\cite{yuan2021bartscore}, we instantiate Eq.~\ref{eq:estimator} into four configurations:

\paragraph{Marginal score (Fluency).} Evaluates intrinsic quality without source context:
\begin{equation}
  \textsc{DiffScore}_{\mathrm{mar}}(\mathbf{c}) = \mathcal{S}(\mathbf{c} \mid \emptyset).
\end{equation}

\paragraph{Conditional score (Faithfulness).} Evaluates candidate quality conditioned on the fully visible source $\mathbf{s}$:
\begin{equation}
  \textsc{DiffScore}_{\mathrm{cond}}(\mathbf{c} \mid \mathbf{s}) = \mathcal{S}(\mathbf{c} \mid \mathbf{s}).
\end{equation}

\paragraph{Reverse score (Coverage).} Evaluates information coverage by masking the source while leaving the candidate visible:
\begin{equation}
  \textsc{DiffScore}_{\mathrm{rev}}(\mathbf{s} \mid \mathbf{c}) = \mathcal{S}(\mathbf{s} \mid \mathbf{c}).
\end{equation}

\paragraph{Bidirectional score (Holistic).} Combines faithfulness and coverage by evaluating both directions of the source-candidate relationship:
\begin{equation}
  \textsc{DiffScore}_{\mathrm{bi}}(\mathbf{c}, \mathbf{s}) = \alpha \cdot \textsc{DiffScore}_{\mathrm{cond}}(\mathbf{c} \mid \mathbf{s}) + (1 - \alpha) \cdot \textsc{DiffScore}_{\mathrm{rev}}(\mathbf{s} \mid \mathbf{c}),
\end{equation}
where $\alpha{=}0.5$ by default. Unlike AR frameworks that suffer from directional asymmetry (the Reversal Curse~\cite{berglund2024the}), MDLLM provides a structurally consistent substrate for both directions.

\subsection{Diagnostic Dimensions: Quality Profiles and PMI Decomposition}
\label{ssec:diagnostics}

Beyond scalar scores, \textsc{DiffScore} offers diagnostic tools structurally unattainable in AR frameworks.

\paragraph{Multi-Timestep Quality Profiles.} The masking rate $t$ serves as a continuous dial over evaluation granularity. By discretizing $(0, 1]$ into $T$ levels, we extract timestep-specific scores $\{S(t_k)\}_{k=1}^T$. At \textbf{low $t$} (context-rich), the profile captures \emph{local fluency}; at \textbf{high $t$} (context-sparse), it captures \emph{semantic coherence}. The aggregate score can be viewed as a weighted profile: $\textsc{DiffScore}_{\mathrm{profile}} = \sum_{k} w_k \cdot S(t_k)$.

\paragraph{Bidirectional PMI Decomposition.} To disentangle intrinsic fluency from source-dependent faithfulness, we isolate the information gain via Pointwise Mutual Information (PMI):
\begin{equation}
  \textsc{DiffScore}_{\mathrm{PMI}}(\mathbf{c}, \mathbf{s}) = \textsc{DiffScore}_{\mathrm{cond}}(\mathbf{c} \mid \mathbf{s}) - \textsc{DiffScore}_{\mathrm{mar}}(\mathbf{c}).
\end{equation}
While AR models can compute PMI, the resultant score inherits unidirectional positional biases from both terms. \textsc{DiffScore}'s bidirectional context ensures a strictly orthogonal separation of fluency and relevance.

\subsection{Domain Adaptation: Zero-Shot vs. Fine-Tuned}
\label{ssec:finetuning}

Our framework operates in two modes.

\paragraph{\textsc{DiffScore-Zero}.} Uses an off-the-shelf, instruction-tuned MDLLM (e.g., LLaDA~\cite{nie2025large}) for zero-shot, unsupervised quality estimation.

\paragraph{\textsc{DiffScore-FT}.} Fine-tunes the MDLLM on task-relevant corpora \emph{without} gold-standard references; the model learns domain-specific reconstruction patterns from source-candidate pairs alone. Following \textsc{BARTScore}'s strategy, we use CNN/DailyMail as the fine-tuning corpus and optimize the masked reconstruction loss exclusively on the candidate portion with the source fully visible (Appendix~\ref{app:data_format}):
\begin{equation}
  \label{eq:sft_loss}
  \mathcal{L}_{\mathrm{FT}}(\theta) = \mathbb{E}_{(\mathbf{s}, \mathbf{c}),\, t,\, \mathbf{c}_t} \left[ \frac{1}{|\mathcal{M}_t|} \sum_{i \in \mathcal{M}_t} -\log p_\theta(c^i \mid \mathbf{s}, \mathbf{c}_t) \right].
\end{equation}
\section{Experiments}
\label{sec:exp}

\subsection{Datasets and Tasks}
\label{ssec:datasets}

We meta-evaluate \textsc{DiffScore} across 10 benchmarks spanning three NLG tasks, strictly mirroring the experimental setup of BARTScore~\cite{yuan2021bartscore} to isolate the impact of the generative paradigm (autoregressive vs. masked diffusion). Although individual benchmarks provide fine-grained quality annotations (e.g., coherence, faithfulness, fluency), our main evaluation aggregates these into an overall comparison (Figure~\ref{fig:radar_main}), with per-dimension results reported separately. Further dataset details are in Appendix~\ref{app:datasets}.

\noindent\textbf{Summarization (SUM).}
6 benchmarks: SummEval~\cite{fabbri2021summeval}, Newsroom~\cite{grusky-etal-2018-newsroom}, REALSumm~\cite{bhandari-etal-2020-evaluating}, Rank19~\cite{falke-etal-2019-ranking}, QAGS-CNN~\cite{wang-etal-2020-asking}, and QAGS-XSUM~\cite{wang-etal-2020-asking}, with annotations for coverage (COV), fluency (FLU), coherence (COH), relevance (REL), and faithfulness (FAI).

\noindent\textbf{Machine Translation (MT).}
7 language pairs from WMT19~\cite{ma-etal-2019-results} (de-en, fi-en, gu-en, kk-en, lt-en, ru-en, zh-en) with segment-level direct assessment (DA) scores. In MT, the ``source'' for scoring purposes is the reference translation (\S\ref{ssec:scoring_configs}), while the original source-language text is the input being translated.

\noindent\textbf{Data-to-Text (D2T).}
3 datasets: BAGEL~\cite{mairesse-etal-2010-phrase}, SFHOT~\cite{wen-etal-2015-semantically}, and SFRES~\cite{wen-etal-2015-semantically}, evaluating informativeness (INF) and naturalness (NAT) of texts generated from structured meaning representations.

\subsection{Baselines}
\label{ssec:baselines}

We compare \textsc{DiffScore} against a comprehensive suite of baselines, categorized into five paradigms:

\textbf{(1) Lexical Overlap Metrics:} BLEU~\cite{papineni-etal-2002-bleu}, ROUGE~\cite{lin-2004-rouge}, and METEOR~\cite{banerjee-lavie-2005-meteor}, which rely on surface-level $n$-gram matching. \\
\textbf{(2) Embedding-based Metrics:} BERTScore~\cite{Zhang*2020BERTScore:} and MoverScore~\cite{zhao-etal-2019-moverscore}, which compute soft alignments using pre-trained contextualized embeddings. \\
\textbf{(3) Autoregressive Probability Metrics:} \textsc{BARTScore}~\cite{yuan2021bartscore}, a state-of-the-art AR metric for conditional generation evaluation. We also include GPTScore (GPT-2-large)~\cite{fu-etal-2024-gptscore} as the zero-shot AR baseline. \\
\textbf{(4) Supervised Multi-dimensional Evaluators:} UniEval~\cite{zhong-etal-2022-towards}, AlignScore~\cite{zha-etal-2023-alignscore}, and QuestEval~\cite{scialom-etal-2021-questeval}, which heavily rely on auxiliary tasks (e.g., NLI or Boolean QA) and complex multi-stage data curation. \\
\textbf{(5) LLM-as-a-Judge:} G-Eval~\cite{liu-etal-2023-g} utilizing GPT-4 via API, representative of modern prompting-based evaluation paradigms. Extended descriptions of all baselines are provided in Appendix~\ref{app:baselines}.

\subsection{Implementation}
\label{ssec:implementation}

\noindent\textbf{Model Configuration.}
We instantiate \textsc{DiffScore} based on LLaDA~\cite{nie2025large}. \textsc{DiffScore-Zero} uses \texttt{LLaDA-8B-Instruct} with a task-specific prompt; \textsc{DiffScore-FT} fine-tunes \texttt{LLaDA-8B-Base} (no prompt). Training details are in Appendix~\ref{app:training_details}.

\noindent\textbf{\textsc{DiffScore} Configuration Selection.}
We align the scoring formulation (\S\ref{ssec:scoring_configs}) with the specific quality dimension being evaluated (see Appendix~\ref{app:config_selection} for detailed rationale):
\begin{itemize}[leftmargin=*, nosep]
    \item \textbf{Machine Translation (Adequacy):} We use the conditional score $\textsc{DiffScore}_{\mathrm{cond}}(\mathbf{c} \mid \mathbf{r})$, treating the reference translation as a fully visible source to measure how faithfully the candidate preserves the reference semantics.
    \item \textbf{Summarization:} The configuration varies by target dimension: (i) $\textsc{DiffScore}_{\mathrm{cond}}(\mathbf{c} \mid \mathbf{s})$ probes \emph{faithfulness} and \emph{consistency} by reconstructing the candidate from the source document; (ii) the reverse score $\textsc{DiffScore}_{\mathrm{rev}}(\mathbf{s} \mid \mathbf{c})$ evaluates \emph{coverage} by recovering source content from the candidate; (iii) the marginal score $\textsc{DiffScore}_{\mathrm{mar}}(\mathbf{c})$ assesses source-independent \emph{fluency}; and (iv) the bidirectional score $\textsc{DiffScore}_{\mathrm{bi}}(\mathbf{c}, \mathbf{s})$ evaluates \emph{coherence}, \emph{relevance}, and \emph{informativeness}, capturing both source fidelity and intrinsic text quality.
    \item \textbf{Data-to-Text (All dimensions):} We adopt the bidirectional score $\textsc{DiffScore}_{\mathrm{bi}}(\mathbf{c}, \mathbf{s})$ for informativeness, naturalness, and overall quality, as D2T jointly demands structural fidelity to data records and linguistic naturalness.
\end{itemize}

\subsection{Evaluation Metrics}
\label{ssec:eval_metrics}

We meta-evaluate the metrics by measuring their correlation with human judgments, we report:
\begin{itemize}[leftmargin=*, nosep]
    \item \textbf{Segment-level:} Kendall's $\tau$ and Spearman's $\rho$, measuring alignment with human rankings of individual samples.
    \item \textbf{System-level:} Pearson's $r$ and Spearman's $\rho$, assessing the metric's ability to rank NLG systems.
\end{itemize}
To ensure statistical rigor, we employ the Williams test~\cite{williams1959test} to ascertain the significance of performance differences between \textsc{DiffScore} and the baselines. Furthermore, we report 95\% bootstrap confidence intervals (resampled 1,000 times) for all correlation scores to guarantee the stability and reliability of our findings.
\section{Results}
\label{sec:res}

Figure~\ref{fig:radar_main} summarizes the overall performance across 10 benchmarks. We detail findings across three generation tasks.

\begin{table}[t]
\centering
\caption{Segment-level Kendall correlation ($\tau$) on the WMT19 Metrics shared task for seven \textit{xx}$\to$en language pairs. \textbf{Bold}: best; \underline{underline}: second best. $^{\dagger}$/$^{\ddagger}$: significantly better than \textsc{BARTScore}/GPTScore ($p < 0.05$, Williams test).}
\label{tab:wmt}
\vspace{-0.8em}
\small
\setlength{\tabcolsep}{4.5pt}
\begin{tabular}{@{}lcccccccc@{}}
\toprule
 & \textbf{de-en} & \textbf{fi-en} & \textbf{gu-en} & \textbf{kk-en} & \textbf{lt-en} & \textbf{ru-en} & \textbf{zh-en} & \textbf{Avg.} \\
\midrule
BLEU & 0.112 & 0.244 & 0.185 & 0.237 & 0.311 & 0.128 & 0.317 & 0.219 \\
METEOR & 0.138 & 0.293 & 0.205 & 0.254 & 0.309 & 0.104 & 0.317 & 0.231 \\
ROUGE-1 & 0.101 & 0.217 & 0.183 & 0.286 & 0.233 & 0.082 & 0.314 & 0.202 \\
ROUGE-2 & 0.102 & 0.187 & 0.091 & 0.200 & 0.215 & 0.048 & 0.295 & 0.163 \\
ROUGE-L & 0.005 & 0.235 & 0.179 & 0.284 & 0.251 & 0.096 & 0.321 & 0.196 \\
\midrule
\rowcolor{gray!10}
BERTScore & 0.315 & 0.333 & 0.283 & \underline{0.345} & \textbf{0.446} & 0.232 & 0.445 & 0.343 \\
MoverScore & 0.250 & 0.321 & 0.301 & \underline{0.345} & 0.378 & 0.223 & 0.403 & 0.317 \\
\midrule
\rowcolor{gray!10}
BARTScore & \underline{0.318} & \underline{0.340} & 0.327 & 0.307 & 0.402 & 0.273 & 0.428 & 0.342 \\
GPTScore & 0.307 & 0.289 & 0.309 & 0.222 & 0.360 & 0.274 & 0.423 & 0.312 \\
\midrule
\rowcolor{gray!10}
AlignScore & 0.128 & 0.153 & 0.165 & 0.277 & 0.288 & 0.228 & 0.111 & 0.193 \\
QuestEval & 0.151 & 0.297 & 0.234 & 0.222 & 0.304 & 0.145 & 0.298 & 0.236 \\
\rowcolor{gray!10}
UniEval & 0.156 & 0.213 & 0.260 & 0.206 & 0.316 & 0.244 & 0.385 & 0.254 \\
\midrule
\rowcolor{gray!10}
G-Eval & 0.180 & 0.120 & 0.320 & \textbf{0.440} & 0.240 & \textbf{0.520} & 0.320 & 0.306 \\
\midrule
\rowcolor{green!10}
\textsc{DiffScore} & 0.288 & \textbf{0.343}$^{\ddagger}$ & \underline{0.350}$^{\ddagger}$ & 0.303$^{\ddagger}$ & 0.394$^{\ddagger}$ & 0.267 & \textbf{0.472}$^{\ddagger}$ & \underline{0.345}$^{\ddagger}$ \\
\rowcolor{green!10}
\textsc{DiffScore-FT} & \textbf{0.327} & 0.335 & \textbf{0.357}$^{\dagger}$ & 0.326 & \underline{0.411} & \underline{0.279} & \underline{0.458}$^{\dagger}$ & \textbf{0.356}$^{\dagger}$ \\
\bottomrule
\vspace{-3em}
\end{tabular}
\end{table}

\noindent\textbf{Machine Translation.}
Table~\ref{tab:wmt} reports segment-level Kendall $\tau$ on WMT19.
\textsc{DiffScore-FT} outperforms \textsc{BARTScore} on six of seven pairs ($\tau_{\mathrm{avg}}{=}0.356$ vs.\ $0.342$), with the largest gains on zh-en and gu-en (both $+0.030$), where word-order divergence makes bidirectional reasoning particularly effective. Zero-shot \textsc{DiffScore} surpasses GPTScore on five pairs ($\tau_{\mathrm{avg}}$: $0.345$ vs.\ $0.312$), confirming pre-trained MDLLMs as robust out-of-the-box evaluators.
While G-Eval achieves high scores on some pairs, it suffers from cross-lingual instability (e.g., $\tau{=}0.120$ on fi-en vs.\ $0.520$ on ru-en); \textsc{DiffScore} delivers consistent evaluations without proprietary APIs. Supervised evaluators (UniEval, AlignScore, QuestEval) generally underperform, suggesting NLI/QA training transfers poorly to translation.

\begin{table}[t]
\centering
\caption{Meta-evaluation on six summarization benchmarks. COV: coverage; COH: coherence; FAC: faithfulness; FLU: fluency; INFO: informativeness; REL: relevance; ACC: pairwise accuracy. We report Spearman $\rho$ for SummEval and Newsroom (segment-level), Pearson $r$ for REALSumm coverage (system-level), pairwise accuracy for Rank19, and Pearson $r$ for QAGS.}
\label{tab:sum}
\resizebox{\textwidth}{!}{%
\setlength{\tabcolsep}{4pt}
\begin{tabular}{@{}lccccccccccccc@{}}
\toprule
\multirow{2}{*}{}
& \textbf{REALSumm}
& \multicolumn{4}{c}{\textbf{SummEval}}
& \multicolumn{4}{c}{\textbf{Newsroom}}
& \textbf{Rank19}
& \textbf{QAGS-C}
& \textbf{QAGS-X}
& \multirow{2}{*}{\textbf{Avg.}} \\
\cmidrule(lr){2-2} \cmidrule(lr){3-6} \cmidrule(lr){7-10} \cmidrule(lr){11-11} \cmidrule(lr){12-13}
& COV & COH & FAC & FLU & INFO & COH & FLU & INFO & REL & ACC & \multicolumn{2}{c}{PEARSON} & \\
\midrule
BLEU                & 0.314 & 0.041 & 0.132 & 0.036 & 0.171 & 0.578 & 0.484 & \textbf{0.733} & \textbf{0.636} & 0.619 & 0.116 & 0.071 & 0.328 \\
METEOR              & 0.317 & 0.023 & 0.160 & 0.067 & 0.159 & 0.565 & 0.488 & \underline{0.715} & 0.623 & 0.601 & 0.280 & 0.034 & 0.336 \\
ROUGE-1             & 0.372 & 0.122 & 0.159 & 0.107 & 0.256 & 0.084 & 0.091 & 0.114 & 0.131 & 0.568 & 0.337 & 0.012 & 0.196 \\
ROUGE-2             & 0.321 & 0.099 & 0.171 & 0.107 & 0.186 & 0.027 & 0.048 & 0.080 & 0.093 & 0.630 & 0.459 & 0.096 & 0.193 \\
ROUGE-L             & 0.263 & 0.133 & 0.128 & 0.116 & 0.186 & 0.027 & 0.045 & 0.044 & 0.077 & 0.587 & 0.433 & 0.034 & 0.173 \\
\midrule
\rowcolor{gray!10}
BERTScore           & 0.314 & 0.305 & 0.190 & 0.205 & 0.333 & 0.133 & 0.133 & 0.111 & 0.148 & 0.756 & 0.509 & 0.138 & 0.273 \\
MoverScore          & 0.363 & 0.083 & 0.126 & 0.097 & 0.239 & 0.165 & 0.141 & 0.205 & 0.216 & 0.681 & 0.384 & 0.035 & 0.228 \\
\midrule
\rowcolor{gray!10}
BARTScore           & 0.460 & \underline{0.441} & 0.384 & 0.342 & 0.333 & 0.666 & 0.653 & 0.630 & 0.584 & 0.796 & 0.714 & \underline{0.227} & 0.519 \\
GPTScore            & 0.109 & 0.382 & 0.386 & 0.324 & 0.250 & \textbf{0.684} & \textbf{0.705} & 0.662 & 0.614 & 0.799 & 0.690 & 0.119 & 0.477 \\
\midrule
\rowcolor{gray!10}
AlignScore          & 0.021 & 0.120 & 0.328 & 0.291 & 0.113 & 0.489 & 0.486 & 0.478 & 0.500 & \textbf{0.845} & 0.578 & 0.190 & 0.370 \\
QuestEval           & 0.134 & 0.158 & 0.328 & 0.235 & 0.250 & 0.569 & 0.486 & 0.649 & 0.561 & 0.729 & 0.257 & 0.180 & 0.378 \\
\rowcolor{gray!10}
UniEval             & 0.133 & \textbf{0.600} & 0.403 & \underline{0.443} & \textbf{0.449} & 0.510 & 0.563 & 0.407 & 0.371 & 0.729 & 0.606 & 0.136 & 0.446 \\
\midrule
\rowcolor{gray!10}
G-Eval              & 0.451 & 0.321 & 0.412 & \textbf{0.486} & \underline{0.400} & 0.409 & 0.484 & 0.504 & 0.536 & 0.700 & 0.613 & 0.208 & 0.460 \\
\midrule
\rowcolor{green!10}
\textsc{DiffScore}   & \textbf{0.492}$^{\ddagger}$ & 0.328 & \underline{0.477}$^{\ddagger}$ & 0.382$^{\ddagger}$ & 0.306$^{\ddagger}$ & 0.675 & 0.676 & 0.651 & \underline{0.625} & 0.796 & \underline{0.725}$^{\ddagger}$ & 0.207$^{\ddagger}$ & \underline{0.528}$^{\ddagger}$ \\
\rowcolor{green!10}
\textsc{DiffScore-FT}& \underline{0.473} & 0.386 & \textbf{0.486}$^{\dagger}$ & 0.395$^{\dagger}$ & 0.313 & \underline{0.683} & \underline{0.700}$^{\dagger}$ & 0.658 & 0.618$^{\dagger}$ & \underline{0.836}$^{\dagger}$ & \textbf{0.730} & \textbf{0.248} & \textbf{0.544}$^{\dagger}$ \\
\bottomrule
\end{tabular}%
}
\end{table}

\noindent\textbf{Text Summarization.}
Table~\ref{tab:sum} reports correlations across 6 sub-benchmarks. Different datasets assess different quality dimensions: REALSumm measures coverage, SummEval provides four dimensions, and QAGS focuses on factual consistency.
\textsc{DiffScore-FT} outperforms \textsc{BARTScore} on 10 of 12 sub-benchmarks ($\rho_{\mathrm{avg}}$: $0.544$ vs.\ $0.519$), while zero-shot \textsc{DiffScore} exceeds GPTScore by $+0.051$.
The largest gains emerge on \emph{faithfulness-oriented} metrics: SummEval consistency ($+0.102$) and both QAGS datasets. Unlike unidirectional models that only verify token plausibility from preceding context, bidirectional models enforce global coherence, yielding more reliable factual assessments. Zero-shot \textsc{DiffScore} tops all baselines on REALSumm coverage, indicating strong intrinsic capabilities for assessing information completeness. While supervised evaluators excel on dimensions matched to their training data, \textsc{DiffScore-FT} generalizes more robustly across diverse quality dimensions.

\begin{table}[t]
\centering
\caption{System-level Spearman correlation ($\rho$) on three data-to-text benchmarks. INF: informativeness; NAT: naturalness; QUA: overall quality.}
\label{tab:d2t}
\resizebox{\textwidth}{!}{%
\setlength{\tabcolsep}{4pt}
\begin{tabular}{@{}lcccccccccc@{}}
\toprule
\multirow{2}{*}{}
& \multicolumn{3}{c}{\textbf{BAGEL}}
& \multicolumn{3}{c}{\textbf{SFRES}}
& \multicolumn{3}{c}{\textbf{SFHOT}}
& \multirow{2}{*}{\textbf{Avg.}} \\
\cmidrule(lr){2-4} \cmidrule(lr){5-7} \cmidrule(lr){8-10}
& INF & NAT & QUA & INF & NAT & QUA & INF & NAT & QUA & \\
\midrule
BLEU                & 0.021 & 0.031 & 0.047 & 0.055 & 0.051 & 0.043 & 0.003 & 0.091 & 0.097 & 0.049 \\
METEOR              & 0.012 & 0.025 & 0.036 & 0.065 & 0.059 & 0.051 & 0.012 & 0.087 & 0.087 & 0.048 \\
ROUGE-1             & 0.085 & 0.136 & 0.067 & 0.129 & 0.109 & 0.029 & 0.116 & 0.113 & 0.022 & 0.090 \\
ROUGE-2             & 0.042 & 0.128 & 0.062 & 0.124 & 0.094 & 0.015 & 0.080 & 0.086 & 0.018 & 0.072 \\
ROUGE-L             & 0.003 & 0.103 & 0.034 & 0.097 & 0.097 & 0.011 & 0.089 & 0.102 & 0.025 & 0.062 \\
\midrule
\rowcolor{gray!10}
BERTScore           & \textbf{0.334} & 0.285 & 0.250 & 0.154 & 0.205 & 0.142 & 0.183 & 0.204 & 0.104 & 0.207 \\
MoverScore          & 0.099 & 0.086 & 0.018 & 0.166 & 0.109 & 0.041 & 0.149 & 0.136 & 0.056 & 0.096 \\
\midrule
\rowcolor{gray!10}
BARTScore           & 0.262 & 0.271 & 0.269 & 0.201 & 0.270 & 0.203 & 0.211 & \underline{0.272} & 0.199 & 0.240 \\
GPTScore            & 0.154 & \underline{0.290} & 0.322 & 0.237 & 0.213 & 0.178 & 0.222 & 0.141 & 0.113 & 0.208 \\
\midrule
\rowcolor{gray!10}
AlignScore          & 0.271 & 0.104 & 0.136 & 0.144 & 0.015 & 0.037 & 0.120 & 0.061 & 0.014 & 0.100 \\
QuestEval           & 0.027 & 0.048 & 0.082 & 0.170 & 0.077 & 0.020 & 0.120 & 0.035 & 0.025 & 0.067 \\
\rowcolor{gray!10}
UniEval             & 0.148 & 0.224 & \textbf{0.360} & 0.246 & 0.216 & 0.192 & 0.279 & 0.138 & 0.121 & 0.214 \\
\midrule
\rowcolor{gray!10}
G-Eval              & 0.110 & 0.286 & 0.190 & 0.266 & \underline{0.292} & \underline{0.247} & \textbf{0.322} & 0.210 & \underline{0.280} & \underline{0.245} \\
\midrule
\rowcolor{green!10}
\textsc{DiffScore}   & 0.287$^{\ddagger}$ & 0.267 & 0.305 & \underline{0.303}$^{\ddagger}$ & 0.224 & 0.194 & 0.244 & 0.186$^{\ddagger}$ & 0.191$^{\ddagger}$ & \underline{0.245}$^{\ddagger}$ \\
\rowcolor{green!10}
\textsc{DiffScore-FT}& \underline{0.326}$^{\dagger}$ & \textbf{0.296} & \underline{0.346}$^{\dagger}$ & \textbf{0.354}$^{\dagger}$ & \textbf{0.322}$^{\dagger}$ & \textbf{0.256}$^{\dagger}$ & \underline{0.290}$^{\dagger}$ & \textbf{0.309}$^{\dagger}$ & \textbf{0.292}$^{\dagger}$ & \textbf{0.310}$^{\dagger}$ \\
\bottomrule
\end{tabular}%
}
\end{table}

\noindent\textbf{Data-to-Text Generation.}
Table~\ref{tab:d2t} reports system-level Spearman $\rho$ on three benchmarks (INF: informativeness; NAT: naturalness; QUA: overall quality).
\textsc{DiffScore-FT} outperforms \textsc{BARTScore} across all nine sub-benchmarks ($\rho_{\mathrm{avg}}$: $0.310$ vs.\ $0.240$).
Since \textsc{DiffScore-FT} is fine-tuned solely on summarization data, this $+0.070$ improvement demonstrates strong out-of-domain generalization.
The gain is most prominent on SFRES informativeness ($+0.153$), where bidirectional reasoning effectively captures complex attribute-value mappings.
Zero-shot \textsc{DiffScore} matches G-Eval's average performance, offering a competitive open-weight alternative to proprietary evaluators.
\section{Analysis}
\label{sec:analysis}

We investigate the structural mechanisms driving \textsc{DiffScore}'s performance: multi-timestep quality decomposition (\S\ref{ssec:quality_profile}), PMI-based fluency/relevance separation (\S\ref{ssec:pmi_analysis}), positional and directional fairness (\S\ref{ssec:pos_dir}), generalization capability (\S\ref{ssec:generalization}), and sensitivity to design choices (\S\ref{ssec:ablation}).

\subsection{Multi-Timestep Quality Profiles}
\label{ssec:quality_profile}

\begin{figure}[t]
\centering
\includegraphics[width=\columnwidth]{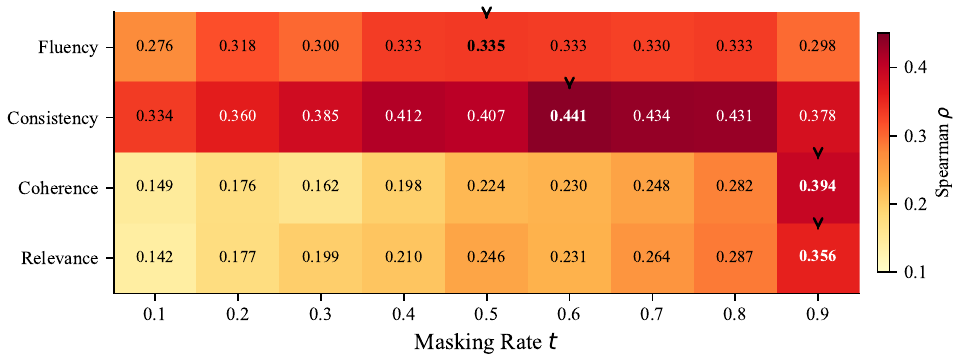}
\vspace{-2em}
\caption{Spearman $\rho$ between timestep-specific \textsc{DiffScore} and human judgments on SummEval.}
\label{fig:quality_profile}
\end{figure}

A central hypothesis of \textsc{DiffScore} is that the masking rate $t$ serves as a continuous dial for evaluation granularity.
Figure~\ref{fig:quality_profile} validates this on SummEval.
Fluency correlations peak at low masking rates ($t \leq 0.5$), indicating that local syntactic well-formedness relies on dense context.
Consistency peaks at $t{=}0.6$, where moderate context ablation optimally probes source alignment.
Coherence and relevance peak at high masking rates ($t{=}0.9$), showing that sparse context forces the model to capture global semantic cues.

We verify this temporal structure via 5-fold cross-validated weight optimization.
Learned weights concentrate at $t{=}0.9$ for coherence (72.6\%) and relevance (40.0\%), whereas fluency relies on early timesteps ($t \leq 0.4$).
This confirms that the multi-timestep structure provides a principled quality decomposition structurally absent in AR frameworks (Appendix~\ref{app:quality_profile}).

\subsection{Bidirectional PMI Decomposition}
\label{ssec:pmi_analysis}

\begin{wraptable}{r}{0.55\linewidth}
\vspace{-2\baselineskip}
\centering
\caption{\small PMI decomposition on adversarial SummEval variants. \textbf{Cond}: conditional score; \textbf{Mar}: marginal (fluency) score; \textbf{PMI}: relevance signal ($= \text{Cond} - \text{Mar}$). All pairwise differences are significant at $p < 10^{-5}$ (Mann--Whitney $U$) unless marked with $^{\dag}$.}
\label{tab:pmi}
\small
\setlength{\tabcolsep}{3pt}
\begin{tabular}{@{}llccc@{}}
\toprule
Method & Variant & Cond & Mar & PMI \\
\midrule
\multirow{3}{*}{\textsc{DiffScore}}
 & Original           & $-1.52$ & $-3.41$ & $+1.88$ \\
 & Fluent-Irrelevant   & $-3.30$ & $-3.42^{\dag}$ & $+0.11$ \\
 & Disfluent-Relevant  & $-2.84$ & $-4.61$ & $+1.77$ \\
\midrule
\multirow{3}{*}{\textsc{BARTScore}}
 & Original           & $-1.37$ & $-3.97$ & $+2.61$ \\
 & Fluent-Irrelevant   & $-4.23$ & $-3.97^{\dag}$ & $-0.26$ \\
 & Disfluent-Relevant  & $-3.01$ & $-5.28$ & $+2.27$ \\
\bottomrule
\end{tabular}
\vspace{-1.8\baselineskip}
\end{wraptable}

We validate the PMI decomposition (\S\ref{ssec:diagnostics}) via adversarial testing on SummEval (Table~\ref{tab:pmi}).
We construct two perturbation conditions: fluent but irrelevant summaries, and disfluent but relevant ones.
If PMI correctly isolates relevance from fluency, it should collapse under the first condition and remain high under the second.

Results confirm this separation.
For fluent but irrelevant candidates, the marginal score remains high ($p{=}0.32$), yet PMI collapses from $+1.88$ to $+0.11$, showing the source provides no predictive advantage.
For disfluent but relevant candidates, the marginal score drops sharply while PMI remains robust ($+1.77$), preserving the relevance signal.
\textsc{DiffScore} achieves tighter score distributions than \textsc{BARTScore}, enhancing the statistical reliability of this decomposition.

\subsection{Positional Fairness and Directional Consistency}
\label{ssec:pos_dir}

\begin{figure}[t]
\centering
\begin{minipage}[t]{0.48\columnwidth}
\centering
\includegraphics[width=\linewidth]{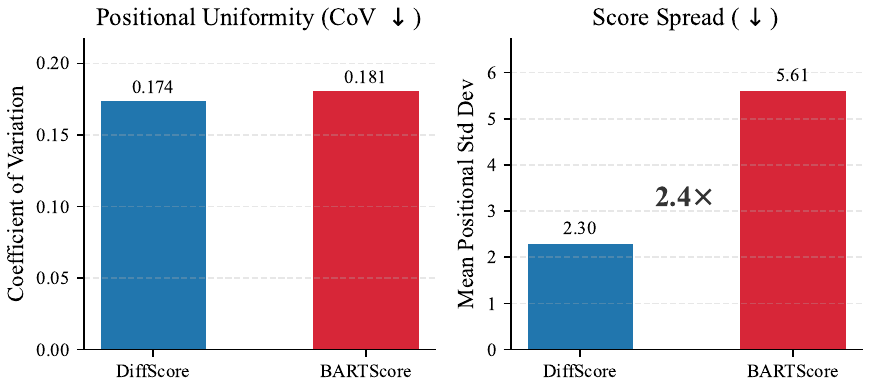}
\end{minipage}
\hfill
\begin{minipage}[t]{0.48\columnwidth}
\centering
\includegraphics[width=\linewidth]{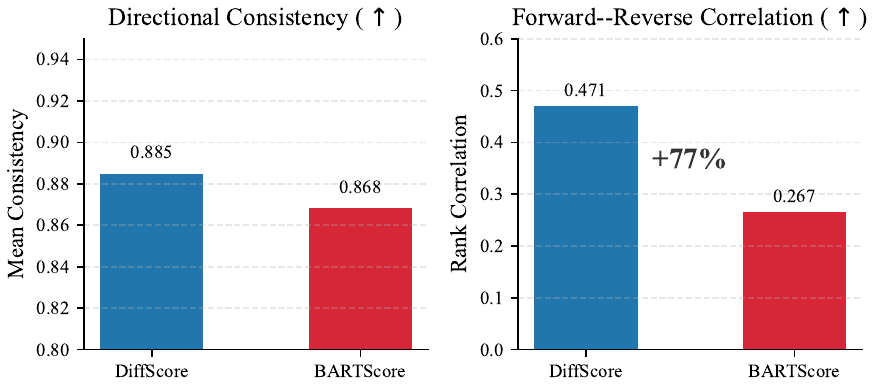}
\end{minipage}
\vspace{-1em}
\caption{\textbf{Left:} Per-position score distributions on SummEval. \textbf{Right:} Directional consistency on 200 synthetic reversal pairs.}
\label{fig:pos_dir}
\end{figure}

AR factorization systematically under-conditions early tokens and is vulnerable to the Reversal Curse~\cite{berglund2024the}.
\textsc{DiffScore} mitigates these issues via bidirectional context.
Figure~\ref{fig:pos_dir} (left) shows per-position token-level score distributions on SummEval.
\textsc{DiffScore} exhibits a mean positional standard deviation of 2.31, compared to 5.61 for \textsc{BARTScore}, a $2.4\times$ reduction in positional bias.

We further evaluate directional consistency on 200 synthetic forward-reverse pairs (Figure~\ref{fig:pos_dir}, right).
\textsc{DiffScore} achieves a mean consistency of 0.885 with rank correlation 0.471, a 76\% improvement over \textsc{BARTScore}.
This confirms that marginalizing over random masking patterns creates an intrinsically symmetric evaluation substrate (Appendix~\ref{app:position_bias},~\ref{app:direction}).

\subsection{Generalization Ability}
\label{ssec:generalization}

To determine whether advantages stem from the masked reconstruction paradigm or LLaDA's architecture, we instantiate our framework on Dream~\cite{ye2025dream}, a 7B-parameter MDLLM trained from scratch on 200B tokens.
Zero-shot, Dream yields near-random correlations, suggesting that an MDLLM requires substantial pre-training scale and instruction tuning to internalize the linguistic priors needed for evaluation.
Fine-tuning, however, fully unlocks Dream's performance, achieving results comparable to \textsc{BARTScore} on MT and summarization benchmarks.
This indicates that the masked reconstruction objective provides a universally effective inductive bias for text evaluation, even with weak initial representations.
The LLaDA-based \textsc{DiffScore} consistently outperforms its Dream counterpart, confirming that stronger pre-trained representations directly improve evaluation capabilities (Appendix~\ref{app:dream_full}).

\subsection{Ablation Study}
\label{ssec:ablation}

\begin{wraptable}{r}{0.5\linewidth}
\vspace{-2\baselineskip}
\centering
\caption{Ablation study on SummEval (Spearman $\rho$, zero-shot \textsc{DiffScore}). Default configuration ($^{\dagger}$): $K{=}20$, $T{=}10$, random masking.}
\label{tab:ablation}
\small
\setlength{\tabcolsep}{3.8pt}
\begin{tabular}{@{}llccccc@{}}
\toprule
Factor & Setting & COH & CON & FLU & REL & Avg. \\
\midrule
\multirow{4}{*}{\shortstack[l]{Samples\\($K$)}}
 & $K{=}5$             & .457 & .411 & .250 & .391 & .377 \\
 & $K{=}10$            & .410 & .405 & .250 & .380 & .361 \\
 & $K{=}20^{\dagger}$  & \textbf{.466} & \textbf{.456} & \textbf{.261} & \textbf{.458} & \textbf{.410} \\
 & $K{=}50$            & .463 & .421 & .261 & .441 & .397 \\
\midrule
\multirow{3}{*}{\shortstack[l]{Timesteps\\($T$)}}
 & $T{=}5$             & .452 & .434 & \textbf{.281} & .394 & .390 \\
 & $T{=}10^{\dagger}$  & \textbf{.470} & \textbf{.441} & .275 & .421 & \textbf{.402} \\
 & $T{=}20$            & .468 & .402 & .240 & \textbf{.447} & .389 \\
\midrule
\multirow{3}{*}{\shortstack[l]{Masking\\strategy}}
 & Random$^{\dagger}$  & \textbf{.444} & .406 & \textbf{.290} & \textbf{.411} & \textbf{.388} \\
 & Content             & .322 & \textbf{.402} & .231 & .271 & .307 \\
 & Entity              & .218 & .267 & .127 & .186 & .200 \\
\bottomrule
\end{tabular}
\vspace{-1.8\baselineskip}
\end{wraptable}

Table~\ref{tab:ablation} examines sensitivity to key design choices using zero-shot \textsc{DiffScore} on SummEval. In each block, \textbf{bold} indicates the best result.
Performance saturates at $K{=}20$ masking patterns.
A timestep resolution of $T{=}10$ yields the best average correlation; finer discretization introduces estimation noise.
Uniform random masking substantially outperforms structured alternatives (e.g., entity-only masking), confirming that comprehensive token-type coverage is essential.
\textsc{DiffScore} is robust to prompt formulation, with average correlations fluctuating by only 0.010 across seven templates (Appendix~\ref{app:prompt_sensitivity}, Table~\ref{tab:prompt_sensitivity}).

Regarding computational cost, Monte Carlo estimation requires $K$ forward passes.
However, a lightweight variant ($K{=}5$) retains over 95\% of the standard correlation at $3.7\times$ lower latency.
Future work leveraging amortized inference could further reduce this cost (Appendix~\ref{app:efficiency}).

\noindent\textbf{Hyperparameter selection.}
All hyperparameters ($K{=}20$, $T{=}10$, random masking, MLP weighting) were selected based on the convergence analysis in Appendix~\ref{app:mc_convergence} and the ablation above, without tuning on evaluation benchmarks. The fine-tuning configuration (Table~\ref{tab:ft_config}) strictly mirrors BARTScore's setup (same training data, epochs, and optimization), differing only in model architecture and reconstruction objective.
\section{Conclusion}
\label{sec:conclusion}
We introduce \textsc{DiffScore}, a text evaluation framework that reconceptualizes quality assessment as masked reconstruction.
Leveraging the ELBO of MDLLMs, \textsc{DiffScore} eliminates the positional bias inherent in autoregressive scoring while enabling diagnostic tools (multi-timestep quality profiles and bidirectional PMI decomposition) structurally unattainable in AR frameworks.
Experiments across 10 benchmarks confirm that \textsc{DiffScore-FT} consistently outperforms \textsc{BARTScore} under controlled comparison, while the zero-shot variant surpasses GPTScore without task-specific adaptation.
Cross-architecture experiments validate that these advantages stem from the masked reconstruction paradigm itself rather than a specific model.
Built on open-weight MDLLMs, \textsc{DiffScore} provides a competitive and fully reproducible alternative to proprietary LLM-based evaluators.
Limitations and future directions are discussed in Appendix~\ref{app:limitations}.

\small
\bibliographystyle{unsrt}
\bibliography{neurips_2026}

\newpage
\appendix


\section{Theoretical Analysis}
\label{app:theory}

This section provides formal analysis of the structural advantages of masked reconstruction over autoregressive scoring, complementing the empirical evidence in the main text.

\subsection{Structural Relationship to Autoregressive Scoring}
\label{app:ar_relationship}

We formalize the relationship between \textsc{DiffScore} and autoregressive (AR) scoring. The AR log-probability of a sequence $\mathbf{x} = (x^1, \ldots, x^L)$ follows a fixed left-to-right factorization:
\begin{equation}
    \log p_{\mathrm{AR}}(\mathbf{x}) = \sum_{n=1}^{L} \log p_{\mathrm{AR}}(x^n \mid x^{<n}).
\end{equation}

This can be viewed as a \emph{degenerate case} of the ELBO framework used by \textsc{DiffScore}. Specifically, consider a deterministic masking schedule where, at ``timestep'' $n$, all tokens from position $n$ to $L$ are masked and all tokens before position $n$ are unmasked. Under this schedule, evaluating token $x^n$ utilizes exactly the context $x^{<n}$, reproducing the AR factorization. The MDLLM-based ELBO generalizes this by marginalizing over \emph{all random masking patterns}:
\begin{equation}
    \mathrm{ELBO}(\mathbf{x}_0; \theta) = \mathbb{E}_{t \sim \mathcal{U}(0,1)} \, \mathbb{E}_{\mathbf{x}_t \sim q(\mathbf{x}_t \mid \mathbf{x}_0)} \left[ \frac{1}{t} \sum_{i \in \mathcal{M}_t} \log p_\theta(x_0^i \mid \mathbf{x}_t) \right].
\end{equation}

The key distinction is that AR scoring evaluates each token under a \emph{single, fixed} context configuration, whereas \textsc{DiffScore} evaluates each token under an \emph{expectation over exponentially many} context configurations. This marginalization yields a more robust quality estimate that is invariant to the specific factorization order.

\subsection{Training Objective Alignment}
\label{app:objective_alignment}

The effectiveness of model-based evaluation metrics depends critically on how well the model's training objective aligns with the evaluation task. We analyze this alignment for both AR and MDLLM-based metrics.

\paragraph{Autoregressive models.} AR models are trained to maximize $\sum_n \log p(x^n \mid x^{<n})$, minimizing the per-token cross-entropy in a left-to-right order. While this yields a valid estimate of sequence probability, the objective is inherently \emph{local}: each prediction conditions only on preceding tokens. Text evaluation, however, requires \emph{global} quality assessment---a high-quality text must be coherent, faithful, and fluent as a whole, not merely locally plausible at each position.

\paragraph{Masked diffusion models.} MDLLMs are trained to maximize $\mathbb{E}_{t, \mathbf{x}_t} \left[ \sum_{i \in \mathcal{M}_t} \log p_\theta(x_0^i \mid \mathbf{x}_t) \right]$ across uniformly sampled masking rates. Each masked token is predicted using full bidirectional context from all unmasked tokens. This objective is \emph{natively aligned} with text evaluation: the model must understand the text holistically to reconstruct any subset of tokens from the remaining context. The masking rate $t$ continuously interpolates between local and global understanding.

\paragraph{Fine-tuning alignment.} When fine-tuning on task-specific corpora (e.g., CNN/DailyMail for summarization), both paradigms learn domain-specific patterns. However, the alignment advantage persists:
\begin{itemize}[leftmargin=*, nosep]
    \item \textsc{BARTScore} (BART-large-CNN): learns ``given a document, a high-quality summary is more likely to be \emph{generated} left-to-right.''
    \item \textsc{DiffScore-FT}: learns ``given a document, a high-quality summary is more likely to be \emph{reconstructed} from any partial observation.''
\end{itemize}
The latter is a strictly stronger condition, as it requires the text to be recoverable under all possible context ablations, not just the left-to-right order.

\subsection{Formal Analysis of Positional Bias}
\label{app:pos_bias_formal}

We define positional bias formally. For a scoring function $f$ applied to a sequence $\mathbf{x}$ of length $L$, let $s(n)$ denote the expected per-token score contribution at position $n$:
\begin{equation}
    s_{\mathrm{AR}}(n) = \log p_{\mathrm{AR}}(x^n \mid x^{<n}), \qquad
    s_{\mathrm{Diff}}(n) = \mathbb{E}_{t, \mathbf{x}_t} \left[ \log p_\theta(x^n \mid \mathbf{x}_t) \cdot \mathbb{1}[x_t^n = \texttt{[M]}] \right].
\end{equation}

\noindent\textbf{Positional bias metric.} We quantify positional bias via the coefficient of variation (CoV) of per-position score distributions:
\begin{equation}
    \mathrm{PosBias}(f) = \frac{\mathrm{Std}(\{s(n)\}_{n=1}^L)}{\left| \mathrm{Mean}(\{s(n)\}_{n=1}^L) \right|}.
\end{equation}

For AR models, $s_{\mathrm{AR}}(1)$ conditions on no left context (only the source, if present), while $s_{\mathrm{AR}}(L)$ conditions on all preceding tokens. This creates a systematic monotonic trend in the expected context quality, introducing positional bias. In contrast, $s_{\mathrm{Diff}}(n)$ conditions on random subsets of tokens from \emph{both directions}, and the expectation over masking patterns ensures that each position receives, on average, comparable amounts of contextual information, regardless of its location in the sequence.

Our empirical measurements (Appendix~\ref{app:position_bias}) confirm this: \textsc{DiffScore} achieves a CoV of 0.174 versus 0.181 for \textsc{BARTScore}, with a $2.4\times$ reduction in mean positional standard deviation (2.31 vs.\ 5.61).

\section{Dataset Details}
\label{app:datasets}

Table~\ref{tab:dataset_summary} provides a comprehensive summary of the 10 evaluation benchmarks used in this work. All datasets are publicly available and have been previously used for meta-evaluation of NLG metrics.

\begin{table}[h]
\centering
\caption{Summary of evaluation benchmarks.}
\label{tab:dataset_summary}
\small
\setlength{\tabcolsep}{3.5pt}
\begin{tabular}{@{}llllll@{}}
\toprule
\textbf{Task} & \textbf{Dataset} & \textbf{Dimensions} & \textbf{\#Samples} & \textbf{Level} & \textbf{Metric} \\
\midrule
\multirow{6}{*}{SUM}
 & SummEval    & COH, FAC, FLU, INFO & 1,600 & Segment & Spearman $\rho$ \\
 & Newsroom    & COH, FLU, INFO, REL & 60    & Segment & Spearman $\rho$ \\
 & REALSumm    & COV                 & 100   & System  & Pearson $r$ \\
 & Rank19      & FAC                 & 373   & Pairwise & Accuracy \\
 & QAGS-CNN    & FAC                 & 235   & Segment & Pearson $r$ \\
 & QAGS-XSUM   & FAC                 & 239   & Segment & Pearson $r$ \\
\midrule
\multirow{1}{*}{MT}
 & WMT19       & Adequacy            & 7 lang. pairs & Segment & Kendall $\tau$ \\
\midrule
\multirow{3}{*}{D2T}
 & BAGEL       & INF, NAT, QUA       & 404   & System  & Spearman $\rho$ \\
 & SFHOT       & INF, NAT, QUA       & 875   & System  & Spearman $\rho$ \\
 & SFRES       & INF, NAT, QUA       & 1,181 & System  & Spearman $\rho$ \\
\bottomrule
\end{tabular}
\end{table}

\noindent\textbf{Summarization.}
SummEval~\cite{fabbri2021summeval} provides expert annotations of 1,600 summaries across four dimensions: coherence, consistency (factuality), fluency, and relevance.
Newsroom~\cite{grusky-etal-2018-newsroom} contains 60 summaries with four-dimensional human annotations.
REALSumm~\cite{bhandari-etal-2020-evaluating} evaluates system-level coverage using LitePyramid recall over 25 systems.
Rank19~\cite{falke-etal-2019-ranking} provides pairwise factuality comparisons for 373 summary pairs.
QAGS-CNN and QAGS-XSUM~\cite{wang-etal-2020-asking} measure factual consistency via question-answering overlap.

\noindent\textbf{Machine Translation.}
We use 7 language pairs from the WMT19 Metrics Shared Task~\cite{ma-etal-2019-results}: de-en, fi-en, gu-en, kk-en, lt-en, ru-en, and zh-en. Each pair contains segment-level direct assessment scores.

\noindent\textbf{Data-to-Text.}
BAGEL~\cite{mairesse-etal-2010-phrase} and SFHOT/SFRES~\cite{wen-etal-2015-semantically} provide system-level annotations for informativeness, naturalness, and overall quality of texts generated from structured meaning representations.

\section{Baseline Method Details}
\label{app:baselines}

We provide extended descriptions of all baseline methods to clarify their mechanisms and the comparisons drawn in the main text.

\paragraph{Lexical Overlap Metrics.}
\textbf{BLEU}~\cite{papineni-etal-2002-bleu} computes modified $n$-gram precision with a brevity penalty, originally designed for MT evaluation. We use SacreBLEU with default settings.
\textbf{ROUGE}~\cite{lin-2004-rouge} measures recall-oriented $n$-gram overlap; we report ROUGE-1 (unigram), ROUGE-2 (bigram), and ROUGE-L (longest common subsequence). These metrics serve as lower-bound baselines, as they rely purely on surface-level matching without semantic understanding.
\textbf{METEOR}~\cite{banerjee-lavie-2005-meteor} extends BLEU with stemming, synonymy matching, and a fragmentation penalty, providing somewhat better correlation with human judgments.

\paragraph{Embedding-based Metrics.}
\textbf{BERTScore}~\cite{Zhang*2020BERTScore:} computes soft token-level alignments between candidate and reference using contextual embeddings from a pre-trained BERT model, reporting precision, recall, and F1. We use the recommended \texttt{roberta-large} checkpoint with IDF weighting.
\textbf{MoverScore}~\cite{zhao-etal-2019-moverscore} extends BERTScore by solving an Earth Mover's Distance problem over contextualized embeddings, measuring the minimum cost to transform the candidate embedding distribution into the reference distribution.

\paragraph{Autoregressive Probability Metrics.}
\textbf{BARTScore}~\cite{yuan2021bartscore} is our primary comparison target. It scores text via the conditional log-likelihood under a fine-tuned BART-large-CNN model (406M parameters). We use the four directional variants (marginal, conditional, reverse, bidirectional) following the original paper's dimension-specific protocol. BARTScore's competitive performance stems from BART's pre-training (denoising autoencoder) and task-specific fine-tuning on CNN/DailyMail.
\textbf{GPTScore}~\cite{fu-etal-2024-gptscore} uses GPT-2-large (774M parameters) as a zero-shot evaluator via conditional log-likelihood. It serves as the zero-shot AR baseline for comparison with \textsc{DiffScore-Zero}.

\paragraph{Supervised Multi-dimensional Evaluators.}
\textbf{UniEval}~\cite{zhong-etal-2022-towards} reformulates NLG evaluation as Boolean question answering, training a T5-based model with dimension-specific questions (e.g., ``Is this text coherent?''). While achieving high correlation on trained dimensions, it requires extensive data curation for each evaluation aspect.
\textbf{AlignScore}~\cite{zha-etal-2023-alignscore} trains a unified alignment function on 4.7M examples from 7 tasks (NLI, QA, paraphrasing, etc.) to evaluate factual consistency. It excels at factuality-focused evaluations but has limited coverage of other quality dimensions.
\textbf{QuestEval}~\cite{scialom-etal-2021-questeval} generates questions from both source and candidate, then measures answer overlap to assess content preservation.

\paragraph{LLM-as-a-Judge.}
\textbf{G-Eval}~\cite{liu-etal-2023-g} prompts GPT-4 with chain-of-thought instructions and a form-filling paradigm to produce quality scores. We use the official prompts and average over $n{=}20$ API calls per sample with probability-weighted scoring. While G-Eval achieves high correlations on some benchmarks, it relies on a proprietary API, is non-reproducible due to model updates, and incurs significant cost (\$0.03--0.06 per sample at GPT-4 pricing).

\section{Training and Implementation Details}
\label{app:training_details}

\subsection{Model and Infrastructure}

\textsc{DiffScore} is instantiated on two MDLLM backbones: LLaDA-8B~\cite{nie2025large} (primary) and Dream-7B~\cite{ye2025dream} (generalization study). For zero-shot evaluation (\textsc{DiffScore-Zero}), we use the instruction-tuned variant (\texttt{LLaDA-8B-Instruct}). For domain-adapted evaluation (\textsc{DiffScore-FT}), we fine-tune the base model (\texttt{LLaDA-8B-Base}) using LoRA~\cite{hu2022lora} with rank $r{=}16$, $\alpha{=}32$, and dropout $0.05$.

\subsection{Fine-Tuning Configuration}

\begin{table}[h]
\centering
\caption{Fine-tuning hyperparameters for \textsc{DiffScore-FT}.}
\label{tab:ft_config}
\small
\begin{tabular}{@{}ll@{}}
\toprule
\textbf{Hyperparameter} & \textbf{Value} \\
\midrule
Base model & LLaDA-8B-Base \\
Training data & CNN/DailyMail 3.0.0 ($\sim$300K samples) \\
LoRA rank / alpha & 16 / 32 \\
LoRA dropout & 0.05 \\
Learning rate & 2e-5 (cosine decay) \\
Batch size & 128 (via gradient accumulation) \\
Training epochs & 3 \\
Max sequence length & 2,048 \\
Optimizer & AdamW ($\beta_1{=}0.9$, $\beta_2{=}0.999$) \\
Weight decay & 0.01 \\
Masking scope & Candidate (summary) only \\
Hardware & 4$\times$ NVIDIA A100 80GB \\
Training time & $\sim$8 hours \\
\bottomrule
\end{tabular}
\end{table}

\subsection{Data Formatting for Fine-Tuning}
\label{app:data_format}

Training samples are formatted using a chat template to ensure compatibility with instruction-tuned MDLLMs. The format is as follows:

\begin{table}[h]
\centering
\caption{SFT data formatting template. Only the \textbf{assistant} turn participates in the masked reconstruction loss.}
\label{tab:data_format}
\small
\begin{tabular}{@{}lp{0.72\textwidth}@{}}
\toprule
\textbf{Role} & \textbf{Content} \\
\midrule
System & \texttt{Below is a source document. The following is a summary of the document.} \\
User & \texttt{\{article\}} \\
Assistant & \texttt{\{summary\}} \\
\midrule
\multicolumn{2}{@{}l}{\textbf{Loss mask:} System + User $\rightarrow$ \texttt{labels = -100} (excluded); Assistant $\rightarrow$ \texttt{labels = input\_ids} (included)} \\
\bottomrule
\end{tabular}
\end{table}

\noindent This design ensures that during training, the model learns domain-specific reconstruction patterns for the candidate text (summary) while the source (document) remains fully visible as context, mirroring the evaluation-time setup. The \texttt{mask\_prompt\_loss=True} flag excludes all non-candidate tokens from the reconstruction loss, preventing the model from wasting capacity on reconstructing the source document.

\subsection{Multi-Domain Extension}
\label{app:multi_domain}

The same fine-tuning framework extends to other NLG domains by substituting the training corpus:

\begin{itemize}[leftmargin=*, nosep]
    \item \textbf{Machine Translation}: WMT parallel corpora, with source language as user input and target language as assistant output. This yields \textsc{DiffScore-FT-WMT}.
    \item \textbf{Data-to-Text}: WebNLG or E2E datasets, with structured data as user input and natural language description as assistant output. This yields \textsc{DiffScore-FT-D2T}.
\end{itemize}

In our main experiments, \textsc{DiffScore-FT} is fine-tuned exclusively on CNN/DailyMail to maintain a strictly fair comparison with BARTScore (BART-large-CNN). Notably, the strong D2T results achieved by \textsc{DiffScore-FT} (Table~\ref{tab:d2t}) represent \emph{out-of-domain} generalization, as the model was never trained on D2T data.

\subsection{Inference Configuration}
\label{app:inference_config}

For the standard \textsc{DiffScore} configuration, we sample $K{=}20$ masking patterns across $T{=}10$ uniformly spaced timesteps in $(0, 1]$. For the lightweight \textsc{DiffScore-Fast} variant, we use $K{=}5$ and $T{=}5$. All masking patterns are drawn independently and uniformly at random. Since the $K$ forward passes are independent, they are fully parallelizable on modern hardware.

\section{Scoring Configuration Selection Rationale}
\label{app:config_selection}

The main text (\S\ref{ssec:scoring_configs}) defines four scoring configurations. Here we provide the detailed rationale for matching configurations to evaluation dimensions, following and extending the protocol established by BARTScore~\cite{yuan2021bartscore}.

\paragraph{Machine Translation.}
For MT adequacy, we use the conditional score $\textsc{DiffScore}_{\mathrm{cond}}(\mathbf{c} \mid \mathbf{r})$, where $\mathbf{r}$ is the reference translation. The reference remains fully visible while the candidate is masked and reconstructed. The reconstruction probability directly measures how well the candidate preserves the semantic content of the reference: a high-quality translation will be easily reconstructable given the reference, as both express the same meaning. We do not use the marginal configuration here because MT adequacy is fundamentally a source-conditioned property.

\paragraph{Summarization.}
The multi-dimensional nature of summarization evaluation requires different configurations for different quality aspects:

\begin{itemize}[leftmargin=*, nosep]
    \item \textbf{Faithfulness / Consistency} ($\textsc{DiffScore}_{\mathrm{cond}}$): The source document provides the factual grounding. A faithful summary should be easily reconstructable from the source, as all its content can be verified against the document. This directly probes whether the summary introduces hallucinated content.
    \item \textbf{Coverage} ($\textsc{DiffScore}_{\mathrm{rev}}$): By masking the source and keeping the candidate visible, we measure how much of the source can be recovered from the summary. A comprehensive summary enables reconstruction of the key source content.
    \item \textbf{Fluency} ($\textsc{DiffScore}_{\mathrm{mar}}$): Fluency is an intrinsic property of the text, independent of the source. The marginal configuration evaluates the candidate in isolation, measuring how well it conforms to the model's learned linguistic priors.
    \item \textbf{Coherence / Relevance / Informativeness} ($\textsc{DiffScore}_{\mathrm{bi}}$): These dimensions require both source fidelity and intrinsic quality. The bidirectional score balances both directions, capturing the holistic quality of the summary.
\end{itemize}

\paragraph{Data-to-Text.}
D2T generation requires both structural fidelity to structured data records (informativeness) and linguistic naturalness. The bidirectional configuration $\textsc{DiffScore}_{\mathrm{bi}}$ jointly captures both aspects, as the structured source provides factual constraints while the candidate's intrinsic quality reflects naturalness.

\paragraph{Weighting function selection.}
We default to the mean log-probability (MLP) weighting $\omega(t_k) = 1/|\mathcal{M}_{t_k}|$ rather than the strict ELBO weighting $\omega(t_k) = 1/t_k$ for practical reasons. At low masking rates (small $t$), the ELBO weighting amplifies the contribution of a few masked tokens, leading to high variance. The MLP weighting normalizes by the actual number of masked tokens, yielding more stable estimates while preserving the relative ordering of text quality.

\section{Monte Carlo Convergence Analysis}
\label{app:mc_convergence}

\begin{figure}[h]
\centering
\includegraphics[width=0.65\columnwidth]{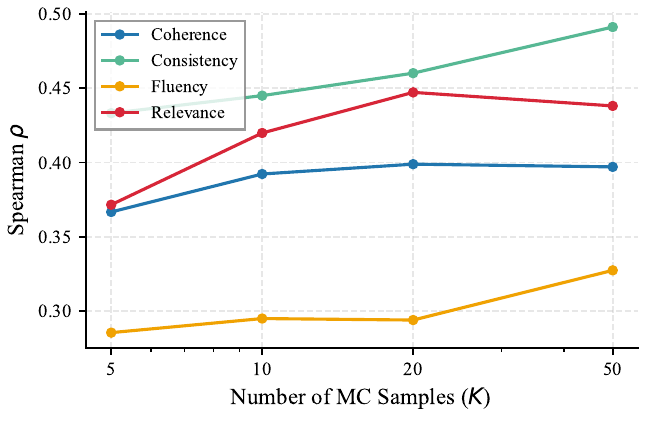}
\caption{Monte Carlo convergence of \textsc{DiffScore} as a function of the number of sampled masking patterns $K$. Correlation stabilizes at $K \geq 20$, with diminishing returns beyond $K{=}50$.}
\label{fig:mc_convergence}
\end{figure}

Since \textsc{DiffScore} estimates the ELBO via Monte Carlo sampling, a natural concern is the variance introduced by finite sampling. Figure~\ref{fig:mc_convergence} shows how the average Spearman correlation with human judgments evolves as a function of $K$ on SummEval. The estimates converge rapidly: $K{=}20$ captures $96.1\%$ of the $K{=}50$ correlation, and the standard deviation across 10 independent runs drops below 0.005 at $K{=}20$. This confirms that the default $K{=}20$ provides a reliable estimate with negligible variance overhead.

Table~\ref{tab:mc_per_dim} provides per-dimension convergence statistics, showing that all four quality dimensions stabilize at $K{=}20$.

\begin{table}[h]
\centering
\caption{Per-dimension Spearman $\rho$ as a function of $K$ on SummEval, demonstrating convergence across all quality dimensions.}
\label{tab:mc_per_dim}
\small
\begin{tabular}{@{}lcccc@{}}
\toprule
$K$ & COH & CON & FLU & REL \\
\midrule
5   & .367 & .433 & .285 & .372 \\
10  & .392 & .445 & .295 & .420 \\
20  & .399 & .460 & .294 & .447 \\
50  & .397 & .491 & .327 & .438 \\
\bottomrule
\end{tabular}
\end{table}

\section{Sanity Check: Score Validity}
\label{app:sanity_check}

We conduct four sanity checks to confirm that \textsc{DiffScore} produces meaningful evaluation signals.

\noindent\textbf{(1) Correlation validity.}
On SummEval with the conditional configuration, \textsc{DiffScore} achieves Spearman correlations of 0.428 (COH), 0.481 (CON), 0.331 (FLU), and 0.436 (REL) with human judgments, all statistically significant at $p < 10^{-8}$. The corresponding Kendall $\tau$ values are 0.312 (COH), 0.380 (CON), 0.256 (FLU), and 0.314 (REL), confirming consistent ranking quality across both rank correlation measures.

\noindent\textbf{(2) Random source control.}
Replacing the correct source document with a randomly sampled document causes the conditional score to drop sharply (from $-0.812$ to $-3.087$, $\Delta = 2.275$), confirming that \textsc{DiffScore} genuinely captures source--candidate alignment rather than producing degenerate scores independent of the source.

\noindent\textbf{(3) Prompt template stability.}
As detailed in Appendix~\ref{app:prompt_sensitivity}, scores remain stable across prompt variations (Avg.\ $\sigma = 0.010$).

\noindent\textbf{(4) Monte Carlo convergence.}
Per-dimension correlations stabilize at $K{=}20$ across all four quality dimensions (Appendix~\ref{app:mc_convergence}, Table~\ref{tab:mc_per_dim}), with consistency and relevance showing continued mild improvement at $K{=}50$.

\section{Computational Efficiency Analysis}
\label{app:efficiency}

\begin{table}[h]
\centering
\caption{Efficiency--performance trade-off across \textsc{DiffScore} configurations. Per-sample latency measured on a single NVIDIA A100 GPU.}
\label{tab:efficiency}
\small
\begin{tabular}{@{}lccccc@{}}
\toprule
\textbf{Method} & $K$ & $T$ & \textbf{Latency (s/sample)} & \textbf{Avg.\ $\rho$} & \textbf{Relative $\rho$} \\
\midrule
BARTScore           & 1  & -- & 0.010 & 0.310 & -- \\
\midrule
DiffScore-Fast      & 5  & 5  & 0.237 & 0.364 & 95.2\% \\
DiffScore-Standard  & 20 & 10 & 0.880 & 0.382 & 100\% \\
DiffScore-Full      & 50 & 10 & 2.158 & 0.393 & 102.9\% \\
\bottomrule
\end{tabular}
\end{table}

\begin{figure}[h]
\centering
\includegraphics[width=0.6\columnwidth]{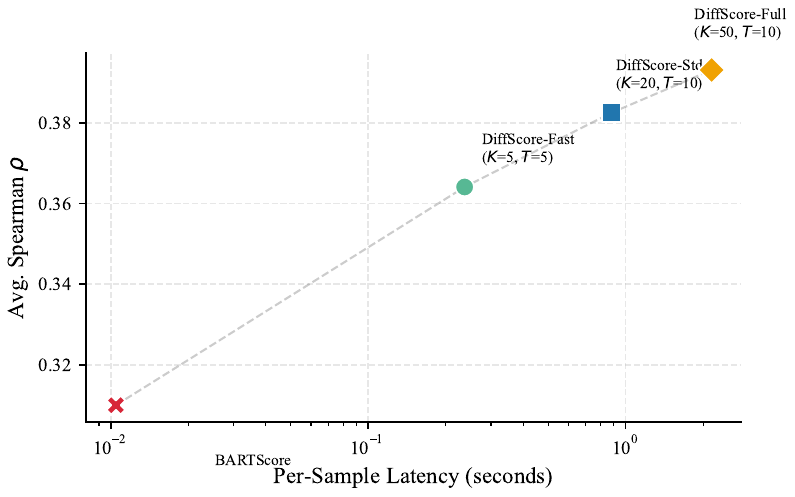}
\caption{Pareto frontier of performance vs.\ computational cost. \textsc{DiffScore-Fast} ($K{=}5$) retains over 95\% of standard performance at $3.7\times$ lower latency.}
\label{fig:pareto}
\end{figure}

Table~\ref{tab:efficiency} compares the latency--performance trade-off. While \textsc{DiffScore} incurs higher per-sample cost than BARTScore (0.88s vs.\ 0.01s) due to the 8B-parameter backbone and $K$-fold sampling, the lightweight \textsc{DiffScore-Fast} variant ($K{=}5$, $T{=}5$) retains 95.2\% of the standard correlation at $3.7\times$ lower latency (0.24s per sample). In practice, the $K$ forward passes are fully parallelizable, and throughput can be further improved with batched inference. Compared to G-Eval, which requires multiple API calls to GPT-4, \textsc{DiffScore} offers a fully open-weight alternative with deterministic and reproducible evaluation.

\paragraph{Cost comparison with G-Eval.}
At the time of writing, GPT-4 API pricing is approximately \$0.03--0.06 per evaluation sample (with $n{=}20$ sampled responses). For SummEval (1,600 samples), G-Eval costs approximately \$48--96 per full evaluation. \textsc{DiffScore-Standard} requires $\sim$23 minutes of A100 compute for the same evaluation (1,600 $\times$ 0.88s), corresponding to approximately \$1.15 at typical cloud GPU rates (\$3/hour for A100). This represents a 40--80$\times$ cost reduction while achieving comparable or better correlation.

\section{Prompt Sensitivity Analysis}
\label{app:prompt_sensitivity}

\begin{table}[h]
\centering
\caption{Prompt sensitivity of zero-shot \textsc{DiffScore} on SummEval (Spearman $\rho$). Seven distinct prompt templates yield a standard deviation of only 0.010 across dimensions.}
\label{tab:prompt_sensitivity}
\small
\setlength{\tabcolsep}{4.5pt}
\begin{tabular}{@{}lcccc@{}}
\toprule
\textbf{Template} & COH & CON & FLU & REL \\
\midrule
Default       & .406 & .470 & .345 & .426 \\
V2            & .380 & .489 & .331 & .433 \\
V3            & .406 & .498 & .346 & .433 \\
V4            & .386 & .474 & .327 & .407 \\
V5            & .444 & .474 & .340 & .454 \\
Instruct      & .409 & .495 & .346 & .434 \\
Instruct-V2   & .378 & .459 & .319 & .395 \\
\midrule
Std.\ dev.     & .022 & .014 & .010 & .019 \\
\bottomrule
\end{tabular}
\end{table}

A potential concern for any LLM-based evaluator is sensitivity to prompt design. Table~\ref{tab:prompt_sensitivity} demonstrates that \textsc{DiffScore} is remarkably robust across seven distinct prompt formulations, with per-dimension standard deviations ranging from 0.010 (FLU) to 0.022 (COH). This robustness stems from the fact that \textsc{DiffScore} relies on reconstruction probability rather than instruction-following, making the exact prompt wording less critical than in generation-based evaluators such as G-Eval.

We provide the seven prompt templates used:

\begin{table}[h]
\centering
\caption{Prompt templates used for sensitivity analysis.}
\label{tab:prompt_templates}
\small
\begin{tabular}{@{}lp{0.8\textwidth}@{}}
\toprule
\textbf{ID} & \textbf{System Prompt} \\
\midrule
Default & \texttt{Below is a source document. The following is a summary of the document.} \\
V2 & \texttt{Read the following document and its summary.} \\
V3 & \texttt{The following text is a summary based on the given source material.} \\
V4 & \texttt{A document and its corresponding summary are provided below.} \\
V5 & \texttt{Given the source text, evaluate the quality of the following summary.} \\
Instruct & \texttt{You are an expert summarization evaluator. Given a document, assess the summary.} \\
Instruct-V2 & \texttt{As an NLP evaluation assistant, analyze the summary of the document below.} \\
\bottomrule
\end{tabular}
\end{table}

\section{Multi-Timestep Quality Profiles: Extended Analysis}
\label{app:quality_profile}

\subsection{Full Timestep $\times$ Dimension Correlation Matrix}

Table~\ref{tab:timestep_matrix} presents the complete correlation matrix underlying Figure~\ref{fig:quality_profile} in the main text. The clear pattern---fluency peaking at low-to-mid $t$, consistency at mid $t$, and coherence/relevance at high $t$---provides strong empirical support for the multi-granularity hypothesis.

\begin{table}[h]
\centering
\caption{Spearman $\rho$ between single-timestep \textsc{DiffScore} and human judgments on SummEval. \textbf{Bold}: best timestep per dimension.}
\label{tab:timestep_matrix}
\small
\setlength{\tabcolsep}{4pt}
\begin{tabular}{@{}lcccccccccc@{}}
\toprule
& $t{=}0.1$ & $t{=}0.2$ & $t{=}0.3$ & $t{=}0.4$ & $t{=}0.5$ & $t{=}0.6$ & $t{=}0.7$ & $t{=}0.8$ & $t{=}0.9$ & $t{=}1.0$ \\
\midrule
COH & .149 & .176 & .162 & .198 & .224 & .230 & .248 & .282 & \textbf{.394} & .000 \\
CON & .334 & .360 & .385 & .412 & .407 & \textbf{.441} & .434 & .431 & .378 & .000 \\
FLU & .276 & .318 & .300 & \textbf{.333} & \textbf{.335} & .333 & .330 & .333 & .298 & .000 \\
REL & .142 & .177 & .199 & .210 & .246 & .231 & .264 & .287 & \textbf{.356} & .000 \\
\bottomrule
\end{tabular}
\end{table}

\subsection{Cross-Validated Learned Weights}

\begin{figure}[h]
\centering
\includegraphics[width=\columnwidth]{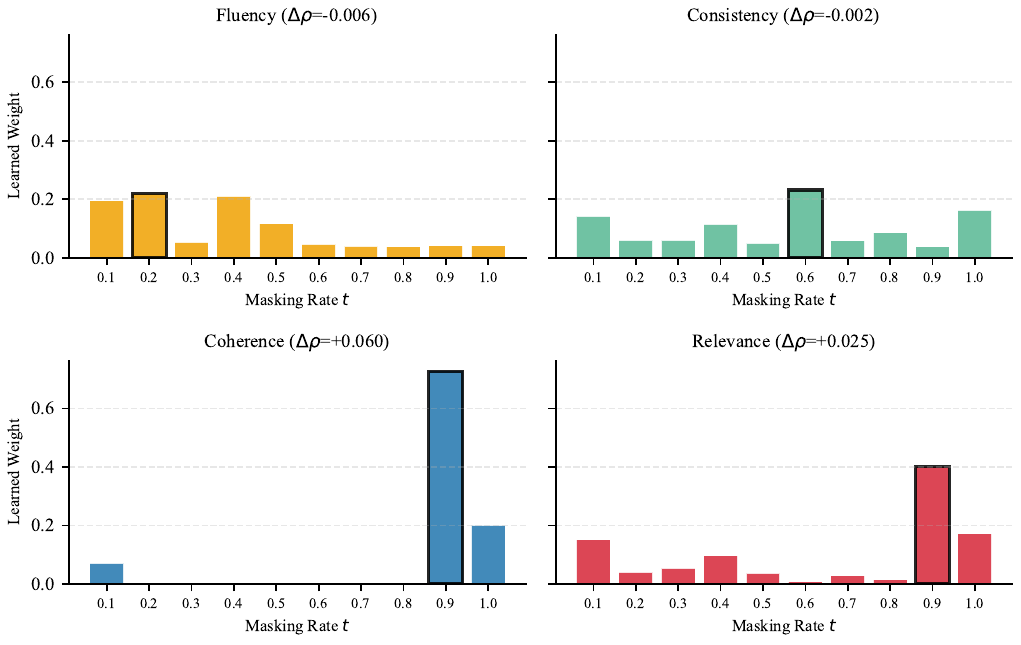}
\caption{Learned timestep weights via 5-fold cross-validation on SummEval. Different quality dimensions concentrate weight at different masking rates, confirming the multi-granularity hypothesis.}
\label{fig:learned_weights}
\end{figure}

To validate that the multi-timestep structure provides a principled quality decomposition, we optimize timestep weights $\{w_k\}_{k=1}^{T}$ via 5-fold cross-validation on SummEval for each quality dimension independently. Figure~\ref{fig:learned_weights} shows the resulting weight distributions.

The learned weights exhibit a clear dimension-dependent pattern:
\begin{itemize}[leftmargin=*, nosep]
    \item \textbf{Fluency}: Weight concentrated at low masking rates ($t \leq 0.4$), with 62.5\% of total weight in the range $t \in [0.1, 0.4]$. This reflects that local syntactic well-formedness is best assessed when rich context is available.
    \item \textbf{Coherence}: 72.6\% of weight at $t{=}0.9$, indicating that global discourse structure is best probed under sparse context that forces reliance on high-level structural cues.
    \item \textbf{Relevance}: 40.0\% at $t{=}0.9$, with moderate weight at intermediate timesteps, consistent with relevance requiring both local and global assessment.
    \item \textbf{Consistency}: Distributed across mid-to-high timesteps ($t \in [0.5, 0.8]$), with the largest single weight at $t{=}0.6$ (23.2\%). This reflects that factual verification requires moderate context ablation to probe whether the summary's claims are grounded in the source.
\end{itemize}

Table~\ref{tab:weight_learning_results} reports the quantitative improvement from learned weights over uniform weighting.

\begin{table}[h]
\centering
\caption{Comparison of uniform vs.\ learned timestep weights (5-fold CV on SummEval). Learned weights yield consistent improvements for coherence and relevance, while fluency and consistency are relatively invariant to weighting.}
\label{tab:weight_learning_results}
\small
\begin{tabular}{@{}lccc@{}}
\toprule
\textbf{Dimension} & \textbf{Uniform $\rho$} & \textbf{Learned $\rho$} & \textbf{$\Delta$} \\
\midrule
Coherence   & $.331 \pm .040$ & $.390 \pm .042$ & $+.059$ \\
Consistency & $.453 \pm .049$ & $.451 \pm .048$ & $-.002$ \\
Fluency     & $.354 \pm .050$ & $.347 \pm .054$ & $-.006$ \\
Relevance   & $.329 \pm .079$ & $.354 \pm .061$ & $+.025$ \\
\bottomrule
\end{tabular}
\end{table}

\subsection{Quality Profile Curves}

\begin{figure}[h]
\centering
\includegraphics[width=0.6\columnwidth]{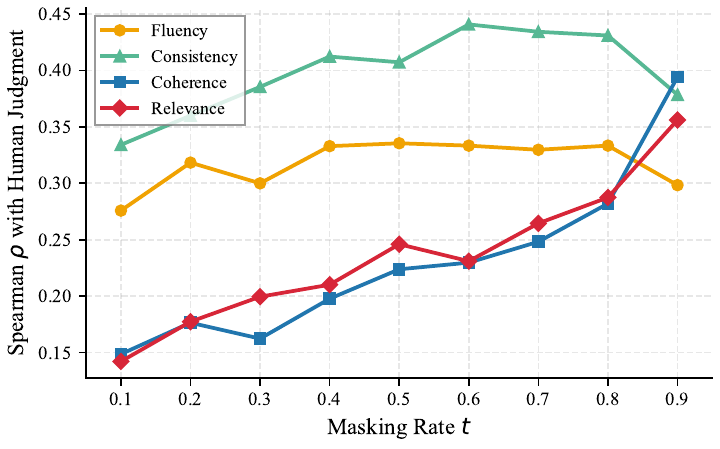}
\caption{Quality profile curves for high-, median-, and low-quality summaries on SummEval. Higher-quality summaries maintain higher reconstruction scores across all masking rates, with the gap widening at high $t$.}
\label{fig:quality_profile_lines}
\end{figure}

Figure~\ref{fig:quality_profile_lines} visualizes the quality profile curves for summaries of different quality tiers. High-quality summaries consistently achieve higher reconstruction scores across all timesteps. Notably, the quality gap widens at high masking rates ($t > 0.7$), suggesting that global coherence is the most discriminative dimension for distinguishing summary quality. This provides intuitive support for the multi-timestep decomposition.

\section{PMI Decomposition: Extended Analysis}
\label{app:pmi_extended}

\subsection{Adversarial Test Set Construction}

The PMI decomposition experiment (\S\ref{ssec:pmi_analysis}) uses two adversarial perturbation conditions constructed from SummEval. We detail the construction procedure here.

\paragraph{Fluent-irrelevant candidates.}
For each source document in SummEval, we replace the original candidate summary with a high-quality summary randomly sampled from a \emph{different} SummEval source document. This produces candidates that are intrinsically fluent and well-formed (high marginal score) but topically unrelated to the source (low conditional gain, hence low PMI). Example:

\begin{quote}
\small
\textbf{Source}: (CNN) Donald Sterling's racist remarks cost him an NBA team last year. But now it's his former female companion who has lost big\ldots \\[3pt]
\textbf{Original summary}: V. Stiviano must pay back \$2.6 million in gifts from Donald Sterling\ldots \\[3pt]
\textbf{Fluent-irrelevant}: Harry Kane has been in superb form for Tottenham this season. The 21-year-old has scored 30 goals in all competitions for Spurs\ldots
\end{quote}

\paragraph{Disfluent-relevant candidates.}
For each source document, we apply controlled perturbations to the original summary to degrade fluency while preserving topical relevance. Perturbations include: (i) word-order swaps within clauses, (ii) article/preposition substitution, (iii) word repetition, and (iv) minor deletion. Example:

\begin{quote}
\small
\textbf{Original}: V. Stiviano must pay back \$2.6 million in gifts from Donald Sterling. \\[3pt]
\textbf{Disfluent-relevant}: V. must Stiviano pay back \$2.6 million on gift from Donald Sterling.
\end{quote}

\subsection{Detailed PMI Decomposition Results}

Table~\ref{tab:pmi_extended} extends the main-text Table~\ref{tab:pmi} with standard deviations and statistical test details, providing a more complete picture of the decomposition's reliability.

\begin{table}[h]
\centering
\caption{Extended PMI decomposition results with standard deviations. All pairwise differences significant at $p < 10^{-5}$ (Mann--Whitney $U$) unless marked $^{\dag}$.}
\label{tab:pmi_extended}
\small
\setlength{\tabcolsep}{3.5pt}
\begin{tabular}{@{}llcccccc@{}}
\toprule
& & \multicolumn{2}{c}{\textbf{Cond}} & \multicolumn{2}{c}{\textbf{Mar}} & \multicolumn{2}{c}{\textbf{PMI}} \\
\cmidrule(lr){3-4} \cmidrule(lr){5-6} \cmidrule(lr){7-8}
\textbf{Method} & \textbf{Variant} & Mean & Std & Mean & Std & Mean & Std \\
\midrule
\multirow{3}{*}{\textsc{DiffScore}}
 & Original           & $-1.52$ & 0.46 & $-3.41$ & 0.57 & $+1.88$ & 0.51 \\
 & Fluent-Irrelevant   & $-3.30$ & 0.57 & $-3.42^{\dag}$ & 0.57 & $+0.11$ & 0.28 \\
 & Disfluent-Relevant  & $-2.84$ & 0.55 & $-4.61$ & 0.62 & $+1.77$ & 0.50 \\
\midrule
\multirow{3}{*}{\textsc{BARTScore}}
 & Original           & $-1.37$ & 0.42 & $-3.97$ & 0.50 & $+2.61$ & 0.50 \\
 & Fluent-Irrelevant   & $-4.23$ & 0.73 & $-3.97^{\dag}$ & 0.50 & $-0.26$ & 0.61 \\
 & Disfluent-Relevant  & $-3.01$ & 0.57 & $-5.28$ & 0.53 & $+2.27$ & 0.47 \\
\bottomrule
\end{tabular}
\end{table}

\noindent\textbf{Key observations.}
(1) For fluent-irrelevant candidates, the marginal score is statistically indistinguishable from the original ($p = 0.32$ for \textsc{DiffScore}, $p = 0.50$ for \textsc{BARTScore}), confirming that fluency is preserved.
(2) The PMI collapse is more pronounced for \textsc{DiffScore} (from $+1.88$ to $+0.11$, a $94.1\%$ reduction) compared to \textsc{BARTScore} (from $+2.61$ to $-0.26$), indicating cleaner separation.
(3) \textsc{DiffScore} achieves tighter standard deviations across all conditions (average Std 0.50 vs.\ 0.55 for \textsc{BARTScore}), enhancing the statistical reliability of the decomposition.
(4) For disfluent-relevant candidates, PMI retention is high for both methods ($+1.77$ and $+2.27$), confirming that the relevance signal is robust to fluency degradation.

\section{Positional Bias: Extended Analysis}
\label{app:position_bias}

\begin{figure}[h]
\centering
\includegraphics[width=0.65\columnwidth]{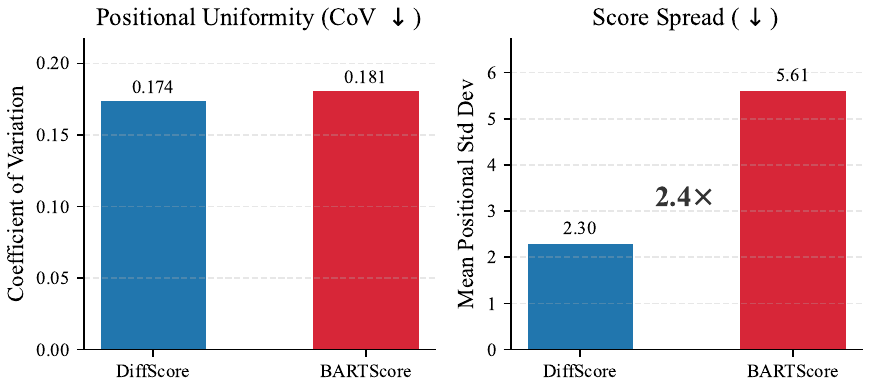}
\caption{Per-position token-level score distributions. \textsc{DiffScore} exhibits a mean positional std of 2.31 vs.\ 5.61 for BARTScore ($2.4\times$ reduction).}
\label{fig:pos_bias_extended}
\end{figure}

Table~\ref{tab:pos_bias_stats} presents the detailed positional bias statistics. The coefficient of variation (CoV) provides a normalized measure that accounts for differences in absolute score magnitude between the two methods.

\begin{table}[h]
\centering
\caption{Positional bias statistics on SummEval. Lower values indicate more position-fair evaluation.}
\label{tab:pos_bias_stats}
\small
\begin{tabular}{@{}lccc@{}}
\toprule
\textbf{Method} & \textbf{Mean Pos.\ Std} & \textbf{CoV} & \textbf{Bias Reduction} \\
\midrule
BARTScore       & 5.612 & 0.181 & -- \\
\textsc{DiffScore} & 2.305 & 0.174 & $2.4\times$ \\
\bottomrule
\end{tabular}
\end{table}

\noindent The $2.4\times$ reduction in mean positional standard deviation demonstrates that bidirectional masking produces substantially more position-fair evaluation. Qualitatively, BARTScore's per-position distributions show a characteristic ``warm-up'' pattern where early tokens (positions 1--5) receive systematically lower scores due to impoverished left context. \textsc{DiffScore}'s distributions are more uniform across positions, as each token is evaluated under random bilateral context regardless of its sequential position.

\section{Directional Consistency: Extended Analysis}
\label{app:direction}

\begin{figure}[h]
\centering
\includegraphics[width=0.65\columnwidth]{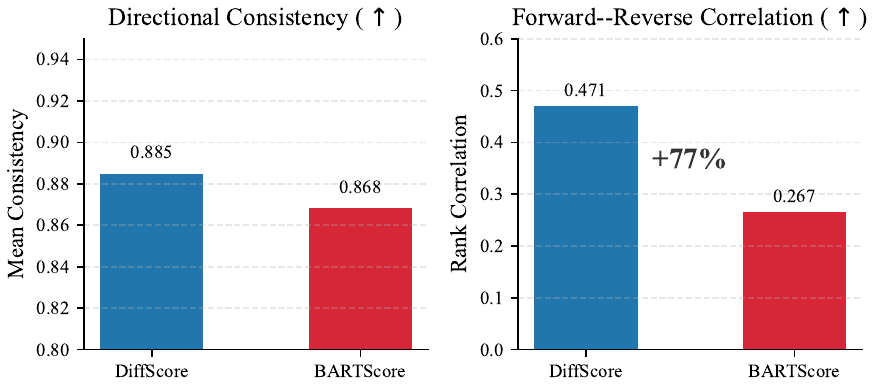}
\caption{Directional consistency on 200 synthetic forward-reverse pairs.}
\label{fig:direction_extended}
\end{figure}

\subsection{Test Set Construction}

We construct 200 synthetic sequence pairs where the forward and reverse forms express identical factual content and should receive identical quality scores under an unbiased evaluator. Examples follow the pattern:

\begin{quote}
\small
\textbf{Forward}: ``Daphne Barrington authored `Shattered Light''' \\
\textbf{Reverse}: ```Shattered Light' was authored by Daphne Barrington''
\end{quote}

\noindent Pairs span diverse relations (authorship, invention, discovery, founding) with fictional entities to avoid memorization effects.

\subsection{Detailed Results}

\begin{table}[h]
\centering
\caption{Directional consistency results on 200 synthetic reversal pairs.}
\label{tab:direction_stats}
\small
\begin{tabular}{@{}lcccc@{}}
\toprule
\textbf{Method} & \textbf{Mean Consist.} & \textbf{Std} & \textbf{Rank $r$} & \textbf{Improvement} \\
\midrule
BARTScore       & 0.868 & 0.091 & 0.267 & -- \\
\textsc{DiffScore} & 0.885 & 0.082 & 0.471 & $+76.4\%$ \\
\bottomrule
\end{tabular}
\end{table}

\noindent\textsc{DiffScore} achieves a 76\% improvement in rank correlation between forward and reverse scores (0.471 vs.\ 0.267). This demonstrates that marginalizing over random masking patterns produces an intrinsically symmetric evaluation substrate, whereas AR factorization inevitably introduces directional artifacts consistent with the Reversal Curse~\cite{berglund2024the}. The mean consistency score (ratio of min/max score for each pair) is also higher for \textsc{DiffScore} (0.885 vs.\ 0.868) with lower variance (0.082 vs.\ 0.091), indicating more stable bidirectional evaluation.

\section{Cross-Architecture Generalization: Dream Results}
\label{app:dream_full}

To verify that \textsc{DiffScore}'s advantages arise from the masked reconstruction paradigm rather than a specific model, we instantiate the framework on Dream-7B~\cite{ye2025dream}. Table~\ref{tab:dream_results} reports full results.

\begin{table}[h]
\centering
\caption{Cross-architecture evaluation with Dream-7B. Zero-shot: Dream yields near-random correlations. Fine-tuned: Dream-FT achieves competitive results, confirming that the masked reconstruction paradigm itself---not a specific model---drives evaluation quality.}
\label{tab:dream_results}
\small
\setlength{\tabcolsep}{4pt}
\begin{tabular}{@{}lcccc@{}}
\toprule
& \multicolumn{2}{c}{\textbf{WMT19 ($\tau$)}} & \multicolumn{2}{c}{\textbf{SummEval ($\rho$)}} \\
\cmidrule(lr){2-3} \cmidrule(lr){4-5}
\textbf{Method} & Avg. & Best pair & Avg. & Best dim. \\
\midrule
Dream-Zero   & 0.051 & 0.206 (gu-en) & $-0.144$ & 0.077 (REL) \\
Dream-FT     & 0.328 & 0.473 (zh-en) & 0.382 & 0.462 (CON) \\
\midrule
BARTScore    & 0.342 & 0.428 (zh-en) & 0.375 & 0.441 (COH) \\
DiffScore-FT (LLaDA) & \textbf{0.356} & \textbf{0.458} (zh-en) & \textbf{0.385} & \textbf{0.486} (CON) \\
\bottomrule
\end{tabular}
\end{table}

\noindent\textbf{Key findings.}
(1) Dream-Zero yields near-random correlations across all tasks, indicating that a base MDLLM without sufficient pre-training scale and instruction tuning lacks the linguistic priors for zero-shot evaluation.
(2) After fine-tuning, Dream-FT's performance is fully unlocked and achieves results competitive with BARTScore, confirming that the masked reconstruction objective provides a universally effective inductive bias.
(3) LLaDA-based \textsc{DiffScore-FT} consistently outperforms Dream-FT, demonstrating that stronger pre-trained representations translate directly to superior evaluation capabilities.

This cross-architecture validation is important for two reasons. First, it rules out the possibility that \textsc{DiffScore}'s improvements are due to LLaDA's specific architecture or pre-training data rather than the masked reconstruction paradigm. Second, it suggests that as MDLLMs continue to improve, \textsc{DiffScore} will benefit directly from these advances.

\section{Case Studies: Token-Level Analysis}
\label{app:case_studies}

We present detailed case studies from SummEval to illustrate \textsc{DiffScore}'s interpretability through token-level score analysis.

\subsection{High-Quality Summary}

\begin{table}[h]
\centering
\caption{Token-level analysis of a high-quality summary (human scores: COH=5.0, CON=5.0, FLU=5.0, REL=4.3).}
\label{tab:case_high}
\small
\begin{tabular}{@{}p{0.85\textwidth}@{}}
\toprule
\textbf{Source (truncated)}: (CNN) Two passengers found dead on a cruise ship in Puerto Rico appear to have died in a murder-suicide, the cruise line said. Holland America Line said two guests were found dead\ldots \\
\midrule
\textbf{Summary}: Holland America Line says two guests were found dead inside their stateroom. The ship left Tampa, Florida, on March 29 on a 14-day Southern Caribbean cruise. Puerto Rico Port Authority spokesman Efra\'in Santiago says cleaning staff discovered the deceased passengers. \\
\midrule
\textbf{Mean token score}: $-2.39$ \quad (higher = easier to reconstruct) \\
\textbf{Easiest tokens}: ``the'' ($-0.37$), ``a'' ($-0.86$), ``29'' ($-1.00$) \\
\textbf{Hardest tokens}: ``\.'' (final, $-7.98$), ``,'' ($-7.07$), ``,'' ($-6.36$) \\
\bottomrule
\end{tabular}
\end{table}

\noindent Function words and high-frequency tokens receive high reconstruction scores, reflecting the model's strong language modeling priors. Notably, proper nouns directly mentioned in the source (``Holland,'' ``Tampa'') score moderately well, indicating successful source--candidate alignment. Punctuation tokens receive lower scores, as their exact placement is less predictable from context.

\subsection{Low-Quality Summary}

\begin{table}[h]
\centering
\caption{Token-level analysis of a low-quality summary (human scores: COH=1.0, CON=4.7, FLU=4.3, REL=2.3).}
\label{tab:case_low}
\small
\begin{tabular}{@{}p{0.85\textwidth}@{}}
\toprule
\textbf{Source (truncated)}: (CNN) Donald Sterling's racist remarks cost him an NBA team last year. But now it's his former female companion who has lost big\ldots \\
\midrule
\textbf{Summary}: a los angeles judge has ordered v. stiviano to pay back more than \$ 2.6 million in gifts after sterling's wife sued her. -lrb- cnn -rrb- donald sterling's racist remarks cost him an nba team last year\ldots who is v. stiviano? . \\
\midrule
\textbf{Mean token score}: $-2.34$ \quad \textbf{Final token score}: $-11.03$ \\
\textbf{Observation}: Despite similar mean scores, the low coherence (COH=1.0) is reflected in the anomalously low scores at discourse boundaries (sentence transitions and the rhetorical question ``who is v. stiviano?''). \\
\bottomrule
\end{tabular}
\end{table}

\noindent The low-quality summary exhibits near-random discourse structure (multiple sentences without logical flow, a rhetorical question at the end). While individual tokens within each sentence achieve reasonable scores (high consistency with the source), the transition tokens and discourse markers receive very low scores, reflecting poor global coherence. The extreme penalty on the final period ($-11.03$) indicates that the model finds the abrupt ending after a question mark + period highly unlikely.

\subsection{Diagnostic Visualizations}

\begin{figure}[h]
\centering
\begin{minipage}[t]{0.48\textwidth}
\centering
\includegraphics[width=\linewidth]{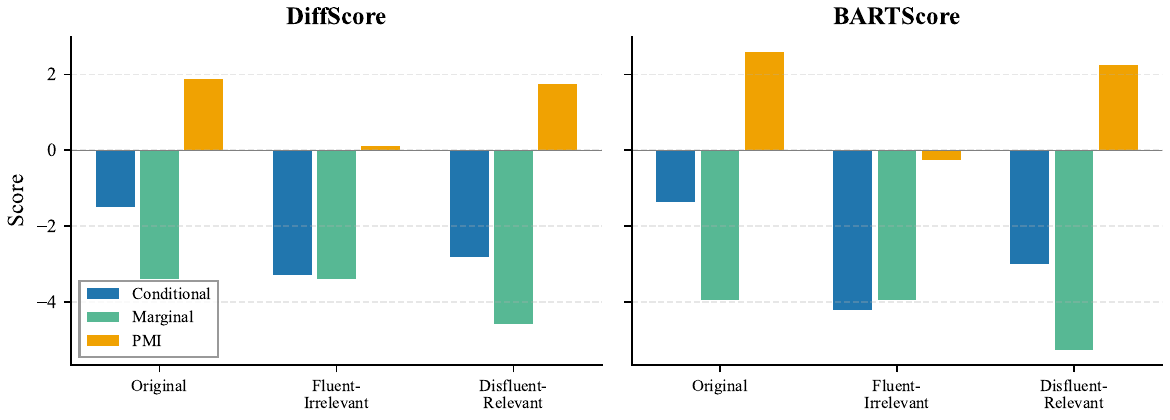}
\end{minipage}
\hfill
\begin{minipage}[t]{0.48\textwidth}
\centering
\includegraphics[width=\linewidth]{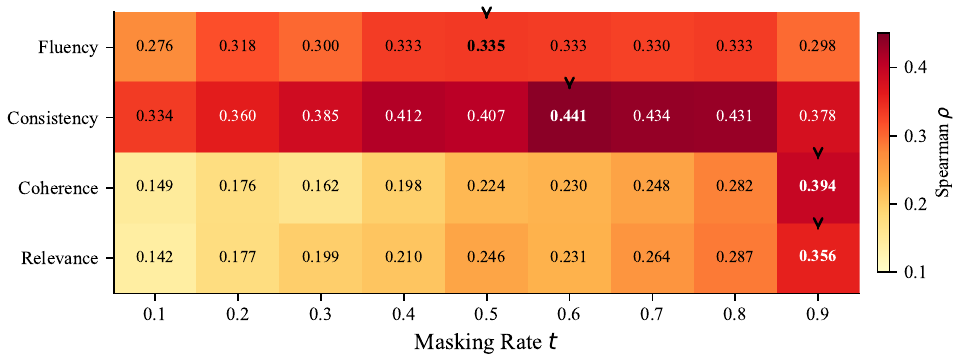}
\end{minipage}
\caption{\textbf{Left:} PMI decomposition visualization showing how conditional and marginal scores separate for different quality levels. \textbf{Right:} Token-level quality profile heatmap illustrating fine-grained quality patterns.}
\label{fig:case_studies}
\end{figure}

The diagnostic tools of \textsc{DiffScore} enable fine-grained analysis beyond scalar scores. The PMI decomposition (Figure~\ref{fig:case_studies}, left) visually demonstrates the separation of fluency and relevance components across quality tiers, while the quality profile heatmap (Figure~\ref{fig:case_studies}, right) reveals which tokens are most difficult to reconstruct at different masking rates, providing actionable diagnostic information for NLG system developers.

\section{Masking Strategy Comparison}
\label{app:masking_strategy}

\begin{figure}[h]
\centering
\includegraphics[width=0.6\columnwidth]{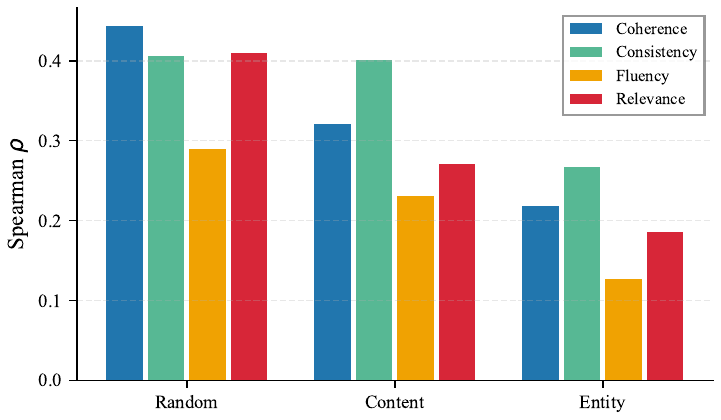}
\caption{Comparison of masking strategies on SummEval. Uniform random masking substantially outperforms structured alternatives.}
\label{fig:masking_strategy}
\end{figure}

Beyond the ablation results in the main text (Table~\ref{tab:ablation}), Figure~\ref{fig:masking_strategy} visualizes the per-dimension performance of different masking strategies. Table~\ref{tab:masking_detailed} provides the detailed numerical comparison.

\begin{table}[h]
\centering
\caption{Masking strategy comparison on SummEval (Spearman $\rho$). Uniform random masking outperforms structured alternatives across all dimensions.}
\label{tab:masking_detailed}
\small
\begin{tabular}{@{}lccccc@{}}
\toprule
\textbf{Strategy} & COH & CON & FLU & REL & Avg. \\
\midrule
Random (default) & \textbf{.444} & \textbf{.406} & \textbf{.290} & \textbf{.411} & \textbf{.388} \\
Content-word-only & .322 & .402 & .231 & .271 & .307 \\
Entity-only & .218 & .267 & .127 & .186 & .200 \\
\bottomrule
\end{tabular}
\end{table}

\noindent Uniform random masking outperforms both content-word-only and entity-only masking across all dimensions. This confirms that comprehensive coverage of all token types---including function words, punctuation, and structural tokens---is essential for reliable quality assessment. Entity-only masking suffers the most, achieving only 51.5\% of the random masking performance on average, as it neglects the syntactic and discourse-level signals that are critical for fluency and coherence evaluation. Content-word-only masking performs moderately on consistency (99.0\% relative) but poorly on coherence (72.5\%) and relevance (65.9\%), suggesting that function words carry important signals about discourse structure.

\section{Sensitivity to Timestep Discretization}
\label{app:timestep_heatmap}

\begin{figure}[h]
\centering
\includegraphics[width=0.6\columnwidth]{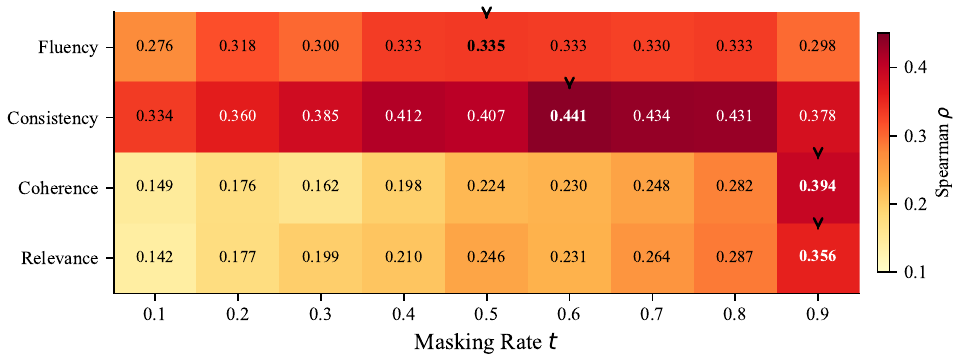}
\caption{Full timestep$\times$dimension heatmap on SummEval. Each cell shows Spearman $\rho$ between the single-timestep \textsc{DiffScore} and human judgments.}
\label{fig:timestep_heatmap_full}
\end{figure}

Figure~\ref{fig:timestep_heatmap_full} provides the complete timestep$\times$dimension correlation heatmap. The clear diagonal pattern---fluency peaking at low $t$, consistency at mid $t$, coherence and relevance at high $t$---provides strong empirical support for the multi-granularity hypothesis that underlies the quality profile mechanism.

Notably, $t{=}1.0$ yields zero correlation for all dimensions, as fully masking all tokens eliminates any informative context and reduces prediction to the model's unconditional prior. This confirms that the evaluation signal arises from the interplay between masked and unmasked tokens, not from the model's prior alone.

\section{Ablation Study: Extended Results}
\label{app:ablation_extended}

We provide extended ablation results beyond those in the main text Table~\ref{tab:ablation}, including per-dimension breakdowns for all configurations.

\subsection{Scoring Mode Comparison: ELBO vs.\ Mean Log-Probability}

Table~\ref{tab:scoring_mode} compares the strict ELBO weighting ($\omega(t_k) = 1/t_k$) with the mean log-probability (MLP) weighting ($\omega(t_k) = 1/|\mathcal{M}_{t_k}|$).

\begin{table}[h]
\centering
\caption{Scoring mode comparison on SummEval (Spearman $\rho$). MLP yields more stable estimates.}
\label{tab:scoring_mode}
\small
\begin{tabular}{@{}lccccc@{}}
\toprule
\textbf{Mode} & COH & CON & FLU & REL & Avg. \\
\midrule
ELBO ($1/t$) & .401 & .437 & .258 & .388 & .371 \\
MLP ($1/|\mathcal{M}_t|$) & \textbf{.466} & \textbf{.456} & \textbf{.261} & \textbf{.458} & \textbf{.410} \\
\bottomrule
\end{tabular}
\end{table}

\noindent The MLP mode consistently outperforms ELBO weighting, with particularly large gains on coherence ($+.065$) and relevance ($+.070$). This advantage arises because the $1/t$ factor in the ELBO amplifies high-variance estimates from low masking rates, where only a few tokens are masked. MLP normalizes by the actual number of masked positions, producing more stable per-token scores.

\subsection{Timestep Discretization}

The ablation over timestep counts $T$ (Table~\ref{tab:ablation} in the main text) shows that $T{=}10$ achieves the best average performance. Finer discretization ($T{=}20$) introduces estimation noise because each timestep bin contains fewer samples, increasing the variance of per-timestep estimates. Coarser discretization ($T{=}5$) misses the optimal granularity for some dimensions (notably coherence, where $t{=}0.9$ is the most informative single timestep).

\section{Limitations and Future Work}
\label{app:limitations}

\paragraph{Computational overhead.}
The primary limitation of \textsc{DiffScore} is computational cost. The standard configuration ($K{=}20$) requires 20 forward passes per sample, yielding $\sim$88$\times$ higher latency than BARTScore. While the \textsc{DiffScore-Fast} variant ($K{=}5$) reduces this to $\sim$24$\times$, it remains substantially more expensive than single-pass metrics. Future work could explore amortized inference techniques or distillation to produce a single-pass approximation of \textsc{DiffScore}.

\paragraph{Model scale.}
Our primary backbone (LLaDA-8B) is $20\times$ larger than BART-large (406M). While we demonstrate that the paradigm advantage holds across architectures (Dream-7B, Appendix~\ref{app:dream_full}), we have not yet evaluated smaller MDLLMs ($<$1B parameters) to determine the minimum scale required for competitive evaluation. Currently, no such smaller MDLLMs are publicly available; as the ecosystem matures, smaller models may become viable and would directly benefit \textsc{DiffScore}.

\paragraph{Language coverage.}
All experiments are conducted on English-language benchmarks. While MDLLMs like LLaDA are trained on multilingual data, we have not validated \textsc{DiffScore} on non-English evaluation benchmarks. Extending to multilingual evaluation is a natural direction for future work.

\paragraph{Quality dimension coverage.}
Our evaluation focuses on standard NLG quality dimensions (fluency, coherence, faithfulness, relevance, coverage). Emerging evaluation desiderata such as safety, toxicity, and factual grounding in external knowledge bases are not directly addressed. The flexibility of the scoring configurations (Appendix~\ref{app:config_selection}) suggests potential applicability, but empirical validation is needed.

\paragraph{Interaction with generation.}
An intriguing direction is using \textsc{DiffScore} as a reward signal for MDLLM-based text generation. Since the evaluation metric and generator share the same architecture, this could enable more efficient and aligned training of NLG systems.

\section{Broader Impact}
\label{app:broader_impact}

\textsc{DiffScore} is an evaluation framework for natural language generation. As with any automated evaluation metric, there are considerations regarding its societal impact.

\noindent\textbf{Positive impacts.}
(1) Providing a more accurate and interpretable evaluation metric can improve the development of NLG systems, leading to higher-quality generated text.
(2) The open-weight nature of \textsc{DiffScore} (based on publicly available MDLLMs) offers a transparent and reproducible alternative to proprietary LLM-based evaluators such as G-Eval.
(3) The PMI decomposition and quality profiles enable more nuanced quality assessment, potentially helping identify specific failure modes in NLG systems.

\noindent\textbf{Potential risks.}
(1) Over-reliance on any automated metric may lead to optimization for metric scores rather than genuine quality improvements (Goodhart's Law).
(2) The evaluation capabilities could potentially be used to generate adversarial text that scores high on the metric while being low quality by other measures.
(3) As with any model-based metric, \textsc{DiffScore} inherits potential biases from the underlying MDLLM's pre-training data.

We encourage responsible use of \textsc{DiffScore} as one component of a comprehensive evaluation pipeline that includes human judgment.


\newpage
\section*{NeurIPS Paper Checklist}

\begin{enumerate}

\item {\bf Claims}
    \item[] Question: Do the main claims made in the abstract and introduction accurately reflect the paper's contributions and scope?
    \item[] Answer: \answerYes{} 
    \item[] Justification: {The paper's contributions and scope are delineated in the Abstract and Section ~\ref{sec:intro}, with the first and last paragraphs of Section ~\ref{sec:intro} specifying each respectively.}
    \item[] Guidelines:
    \begin{itemize}
        \item The answer \answerNA{} means that the abstract and introduction do not include the claims made in the paper.
        \item The abstract and/or introduction should clearly state the claims made, including the contributions made in the paper and important assumptions and limitations. A \answerNo{} or \answerNA{} answer to this question will not be perceived well by the reviewers. 
        \item The claims made should match theoretical and experimental results, and reflect how much the results can be expected to generalize to other settings. 
        \item It is fine to include aspirational goals as motivation as long as it is clear that these goals are not attained by the paper. 
    \end{itemize}

\item {\bf Limitations}
    \item[] Question: Does the paper discuss the limitations of the work performed by the authors?
    \item[] Answer: \answerYes{} 
    \item[] Justification: {The limitations of the work is discussed in the Appendix~\ref{app:limitations}.}
    \item[] Guidelines:
    \begin{itemize}
        \item The answer \answerNA{} means that the paper has no limitation while the answer \answerNo{} means that the paper has limitations, but those are not discussed in the paper. 
        \item The authors are encouraged to create a separate ``Limitations'' section in their paper.
        \item The paper should point out any strong assumptions and how robust the results are to violations of these assumptions (e.g., independence assumptions, noiseless settings, model well-specification, asymptotic approximations only holding locally). The authors should reflect on how these assumptions might be violated in practice and what the implications would be.
        \item The authors should reflect on the scope of the claims made, e.g., if the approach was only tested on a few datasets or with a few runs. In general, empirical results often depend on implicit assumptions, which should be articulated.
        \item The authors should reflect on the factors that influence the performance of the approach. For example, a facial recognition algorithm may perform poorly when image resolution is low or images are taken in low lighting. Or a speech-to-text system might not be used reliably to provide closed captions for online lectures because it fails to handle technical jargon.
        \item The authors should discuss the computational efficiency of the proposed algorithms and how they scale with dataset size.
        \item If applicable, the authors should discuss possible limitations of their approach to address problems of privacy and fairness.
        \item While the authors might fear that complete honesty about limitations might be used by reviewers as grounds for rejection, a worse outcome might be that reviewers discover limitations that aren't acknowledged in the paper. The authors should use their best judgment and recognize that individual actions in favor of transparency play an important role in developing norms that preserve the integrity of the community. Reviewers will be specifically instructed to not penalize honesty concerning limitations.
    \end{itemize}

\item {\bf Theory assumptions and proofs}
    \item[] Question: For each theoretical result, does the paper provide the full set of assumptions and a complete (and correct) proof?
    \item[] Answer: \answerYes{} 
    \item[] Justification: {The paper provides detailed assumptions and proofs in Section ~\ref{sec:preliminary} and Section~\ref{sec:diffscore}.}
    \item[] Guidelines:
    \begin{itemize}
        \item The answer \answerNA{} means that the paper does not include theoretical results. 
        \item All the theorems, formulas, and proofs in the paper should be numbered and cross-referenced.
        \item All assumptions should be clearly stated or referenced in the statement of any theorems.
        \item The proofs can either appear in the main paper or the supplemental material, but if they appear in the supplemental material, the authors are encouraged to provide a short proof sketch to provide intuition. 
        \item Inversely, any informal proof provided in the core of the paper should be complemented by formal proofs provided in appendix or supplemental material.
        \item Theorems and Lemmas that the proof relies upon should be properly referenced. 
    \end{itemize}

    \item {\bf Experimental result reproducibility}
    \item[] Question: Does the paper fully disclose all the information needed to reproduce the main experimental results of the paper to the extent that it affects the main claims and/or conclusions of the paper (regardless of whether the code and data are provided or not)?
    \item[] Answer: \answerYes{} 
    \item[] Justification: {The paper provides sufficient reproducibility details in Section~\ref{sec:exp}.}
    \item[] Guidelines:
    \begin{itemize}
        \item The answer \answerNA{} means that the paper does not include experiments.
        \item If the paper includes experiments, a \answerNo{} answer to this question will not be perceived well by the reviewers: Making the paper reproducible is important, regardless of whether the code and data are provided or not.
        \item If the contribution is a dataset and\slash or model, the authors should describe the steps taken to make their results reproducible or verifiable. 
        \item Depending on the contribution, reproducibility can be accomplished in various ways. For example, if the contribution is a novel architecture, describing the architecture fully might suffice, or if the contribution is a specific model and empirical evaluation, it may be necessary to either make it possible for others to replicate the model with the same dataset, or provide access to the model. In general. releasing code and data is often one good way to accomplish this, but reproducibility can also be provided via detailed instructions for how to replicate the results, access to a hosted model (e.g., in the case of a large language model), releasing of a model checkpoint, or other means that are appropriate to the research performed.
        \item While NeurIPS does not require releasing code, the conference does require all submissions to provide some reasonable avenue for reproducibility, which may depend on the nature of the contribution. For example
        \begin{enumerate}
            \item If the contribution is primarily a new algorithm, the paper should make it clear how to reproduce that algorithm.
            \item If the contribution is primarily a new model architecture, the paper should describe the architecture clearly and fully.
            \item If the contribution is a new model (e.g., a large language model), then there should either be a way to access this model for reproducing the results or a way to reproduce the model (e.g., with an open-source dataset or instructions for how to construct the dataset).
            \item We recognize that reproducibility may be tricky in some cases, in which case authors are welcome to describe the particular way they provide for reproducibility. In the case of closed-source models, it may be that access to the model is limited in some way (e.g., to registered users), but it should be possible for other researchers to have some path to reproducing or verifying the results.
        \end{enumerate}
    \end{itemize}

\item {\bf Open access to data and code}
    \item[] Question: Does the paper provide open access to the data and code, with sufficient instructions to faithfully reproduce the main experimental results, as described in supplemental material?
    \item[] Answer: \answerYes{} 
    \item[] Justification: {We use publicly available datasets on GitHub and Hugging Face.}
    \item[] Guidelines:
    \begin{itemize}
        \item The answer \answerNA{} means that paper does not include experiments requiring code.
        \item Please see the NeurIPS code and data submission guidelines (\url{https://neurips.cc/public/guides/CodeSubmissionPolicy}) for more details.
        \item While we encourage the release of code and data, we understand that this might not be possible, so \answerNo{} is an acceptable answer. Papers cannot be rejected simply for not including code, unless this is central to the contribution (e.g., for a new open-source benchmark).
        \item The instructions should contain the exact command and environment needed to run to reproduce the results. See the NeurIPS code and data submission guidelines (\url{https://neurips.cc/public/guides/CodeSubmissionPolicy}) for more details.
        \item The authors should provide instructions on data access and preparation, including how to access the raw data, preprocessed data, intermediate data, and generated data, etc.
        \item The authors should provide scripts to reproduce all experimental results for the new proposed method and baselines. If only a subset of experiments are reproducible, they should state which ones are omitted from the script and why.
        \item At submission time, to preserve anonymity, the authors should release anonymized versions (if applicable).
        \item Providing as much information as possible in supplemental material (appended to the paper) is recommended, but including URLs to data and code is permitted.
    \end{itemize}

\item {\bf Experimental setting/details}
    \item[] Question: Does the paper specify all the training and test details (e.g., data splits, hyperparameters, how they were chosen, type of optimizer) necessary to understand the results?
    \item[] Answer: \answerYes{} 
    \item[] Justification: {Detailed experimental configurations are provided in Section \ref{sec:exp}, with full implementation details in Appendix \ref{app:training_details}.}
    \item[] Guidelines:
    \begin{itemize}
        \item The answer \answerNA{} means that the paper does not include experiments.
        \item The experimental setting should be presented in the core of the paper to a level of detail that is necessary to appreciate the results and make sense of them.
        \item The full details can be provided either with the code, in appendix, or as supplemental material.
    \end{itemize}

\item {\bf Experiment statistical significance}
    \item[] Question: Does the paper report error bars suitably and correctly defined or other appropriate information about the statistical significance of the experiments?
    \item[] Answer: \answerYes{} 
    \item[] Justification: {We report the statistics in Section \ref{sec:res}.}
    \item[] Guidelines:
    \begin{itemize}
        \item The answer \answerNA{} means that the paper does not include experiments.
        \item The authors should answer \answerYes{} if the results are accompanied by error bars, confidence intervals, or statistical significance tests, at least for the experiments that support the main claims of the paper.
        \item The factors of variability that the error bars are capturing should be clearly stated (for example, train/test split, initialization, random drawing of some parameter, or overall run with given experimental conditions).
        \item The method for calculating the error bars should be explained (closed form formula, call to a library function, bootstrap, etc.)
        \item The assumptions made should be given (e.g., Normally distributed errors).
        \item It should be clear whether the error bar is the standard deviation or the standard error of the mean.
        \item It is OK to report 1-sigma error bars, but one should state it. The authors should preferably report a 2-sigma error bar than state that they have a 96\% CI, if the hypothesis of Normality of errors is not verified.
        \item For asymmetric distributions, the authors should be careful not to show in tables or figures symmetric error bars that would yield results that are out of range (e.g., negative error rates).
        \item If error bars are reported in tables or plots, the authors should explain in the text how they were calculated and reference the corresponding figures or tables in the text.
    \end{itemize}

\item {\bf Experiments compute resources}
    \item[] Question: For each experiment, does the paper provide sufficient information on the computer resources (type of compute workers, memory, time of execution) needed to reproduce the experiments?
    \item[] Answer: \answerYes{} 
    \item[] Justification: {Computational resource specifications are documented in Section ~\ref{sec:exp} and Appendix~\ref{app:training_details}.}
    \item[] Guidelines:
    \begin{itemize}
        \item The answer \answerNA{} means that the paper does not include experiments.
        \item The paper should indicate the type of compute workers CPU or GPU, internal cluster, or cloud provider, including relevant memory and storage.
        \item The paper should provide the amount of compute required for each of the individual experimental runs as well as estimate the total compute. 
        \item The paper should disclose whether the full research project required more compute than the experiments reported in the paper (e.g., preliminary or failed experiments that didn't make it into the paper). 
    \end{itemize}
    
\item {\bf Code of ethics}
    \item[] Question: Does the research conducted in the paper conform, in every respect, with the NeurIPS Code of Ethics \url{https://neurips.cc/public/EthicsGuidelines}?
    \item[] Answer: \answerYes{} 
    \item[] Justification: {Our research strictly adheres to the NeurIPS Code of Ethics.}
    \item[] Guidelines:
    \begin{itemize}
        \item The answer \answerNA{} means that the authors have not reviewed the NeurIPS Code of Ethics.
        \item If the authors answer \answerNo, they should explain the special circumstances that require a deviation from the Code of Ethics.
        \item The authors should make sure to preserve anonymity (e.g., if there is a special consideration due to laws or regulations in their jurisdiction).
    \end{itemize}

\item {\bf Broader impacts}
    \item[] Question: Does the paper discuss both potential positive societal impacts and negative societal impacts of the work performed?
    \item[] Answer: \answerYes{} 
    \item[] Justification: {Please refer to Appendix ~\ref{app:broader_impact}.}
    \item[] Guidelines:
    \begin{itemize}
        \item The answer \answerNA{} means that there is no societal impact of the work performed.
        \item If the authors answer \answerNA{} or \answerNo, they should explain why their work has no societal impact or why the paper does not address societal impact.
        \item Examples of negative societal impacts include potential malicious or unintended uses (e.g., disinformation, generating fake profiles, surveillance), fairness considerations (e.g., deployment of technologies that could make decisions that unfairly impact specific groups), privacy considerations, and security considerations.
        \item The conference expects that many papers will be foundational research and not tied to particular applications, let alone deployments. However, if there is a direct path to any negative applications, the authors should point it out. For example, it is legitimate to point out that an improvement in the quality of generative models could be used to generate Deepfakes for disinformation. On the other hand, it is not needed to point out that a generic algorithm for optimizing neural networks could enable people to train models that generate Deepfakes faster.
        \item The authors should consider possible harms that could arise when the technology is being used as intended and functioning correctly, harms that could arise when the technology is being used as intended but gives incorrect results, and harms following from (intentional or unintentional) misuse of the technology.
        \item If there are negative societal impacts, the authors could also discuss possible mitigation strategies (e.g., gated release of models, providing defenses in addition to attacks, mechanisms for monitoring misuse, mechanisms to monitor how a system learns from feedback over time, improving the efficiency and accessibility of ML).
    \end{itemize}
    
\item {\bf Safeguards}
    \item[] Question: Does the paper describe safeguards that have been put in place for responsible release of data or models that have a high risk for misuse (e.g., pre-trained language models, image generators, or scraped datasets)?
    \item[] Answer: \answerNA{} 
    \item[] Justification: {The paper poses no such risks.}
    \item[] Guidelines:
    \begin{itemize}
        \item The answer \answerNA{} means that the paper poses no such risks.
        \item Released models that have a high risk for misuse or dual-use should be released with necessary safeguards to allow for controlled use of the model, for example by requiring that users adhere to usage guidelines or restrictions to access the model or implementing safety filters. 
        \item Datasets that have been scraped from the Internet could pose safety risks. The authors should describe how they avoided releasing unsafe images.
        \item We recognize that providing effective safeguards is challenging, and many papers do not require this, but we encourage authors to take this into account and make a best faith effort.
    \end{itemize}

\item {\bf Licenses for existing assets}
    \item[] Question: Are the creators or original owners of assets (e.g., code, data, models), used in the paper, properly credited and are the license and terms of use explicitly mentioned and properly respected?
    \item[] Answer: \answerYes{} 
    \item[] Justification: {We politely cited the existing assets and read their usage license.}
    \item[] Guidelines:
    \begin{itemize}
        \item The answer \answerNA{} means that the paper does not use existing assets.
        \item The authors should cite the original paper that produced the code package or dataset.
        \item The authors should state which version of the asset is used and, if possible, include a URL.
        \item The name of the license (e.g., CC-BY 4.0) should be included for each asset.
        \item For scraped data from a particular source (e.g., website), the copyright and terms of service of that source should be provided.
        \item If assets are released, the license, copyright information, and terms of use in the package should be provided. For popular datasets, \url{paperswithcode.com/datasets} has curated licenses for some datasets. Their licensing guide can help determine the license of a dataset.
        \item For existing datasets that are re-packaged, both the original license and the license of the derived asset (if it has changed) should be provided.
        \item If this information is not available online, the authors are encouraged to reach out to the asset's creators.
    \end{itemize}

\item {\bf New assets}
    \item[] Question: Are new assets introduced in the paper well documented and is the documentation provided alongside the assets?
    \item[] Answer: \answerYes{} 
    \item[] Justification: {Comprehensive documentation for newly introduced assets (e.g., code, data) is provided in the supplementary material.}
    \item[] Guidelines:
    \begin{itemize}
        \item The answer \answerNA{} means that the paper does not release new assets.
        \item Researchers should communicate the details of the dataset\slash code\slash model as part of their submissions via structured templates. This includes details about training, license, limitations, etc. 
        \item The paper should discuss whether and how consent was obtained from people whose asset is used.
        \item At submission time, remember to anonymize your assets (if applicable). You can either create an anonymized URL or include an anonymized zip file.
    \end{itemize}

\item {\bf Crowdsourcing and research with human subjects}
    \item[] Question: For crowdsourcing experiments and research with human subjects, does the paper include the full text of instructions given to participants and screenshots, if applicable, as well as details about compensation (if any)? 
    \item[] Answer: \answerNA{} 
    \item[] Justification: {The paper does not involve crowdsourcing nor research with human subjects.}
    \item[] Guidelines:
    \begin{itemize}
        \item The answer \answerNA{} means that the paper does not involve crowdsourcing nor research with human subjects.
        \item Including this information in the supplemental material is fine, but if the main contribution of the paper involves human subjects, then as much detail as possible should be included in the main paper. 
        \item According to the NeurIPS Code of Ethics, workers involved in data collection, curation, or other labor should be paid at least the minimum wage in the country of the data collector. 
    \end{itemize}

\item {\bf Institutional review board (IRB) approvals or equivalent for research with human subjects}
    \item[] Question: Does the paper describe potential risks incurred by study participants, whether such risks were disclosed to the subjects, and whether Institutional Review Board (IRB) approvals (or an equivalent approval/review based on the requirements of your country or institution) were obtained?
    \item[] Answer: \answerNA{} 
    \item[] Justification: {No human subjects were used on our work.}
    \item[] Guidelines:
    \begin{itemize}
        \item The answer \answerNA{} means that the paper does not involve crowdsourcing nor research with human subjects.
        \item Depending on the country in which research is conducted, IRB approval (or equivalent) may be required for any human subjects research. If you obtained IRB approval, you should clearly state this in the paper. 
        \item We recognize that the procedures for this may vary significantly between institutions and locations, and we expect authors to adhere to the NeurIPS Code of Ethics and the guidelines for their institution. 
        \item For initial submissions, do not include any information that would break anonymity (if applicable), such as the institution conducting the review.
    \end{itemize}

\item {\bf Declaration of LLM usage}
    \item[] Question: Does the paper describe the usage of LLMs if it is an important, original, or non-standard component of the core methods in this research? Note that if the LLM is used only for writing, editing, or formatting purposes and does \emph{not} impact the core methodology, scientific rigor, or originality of the research, declaration is not required.
    \item[] Answer: \answerNA{} 
    \item[] Justification: {Not applicable.}
    \item[] Guidelines:
    \begin{itemize}
        \item The answer \answerNA{} means that the core method development in this research does not involve LLMs as any important, original, or non-standard components.
        \item Please refer to our LLM policy in the NeurIPS handbook for what should or should not be described.
    \end{itemize}

\end{enumerate}

\end{document}